\documentclass[sigconf]{acmart}

\usepackage[utf8]{inputenc}    
\usepackage[T1]{fontenc}       
\usepackage{booktabs}
\usepackage{multirow}
\usepackage{subcaption}
\usepackage{pifont}
\usepackage{algorithm}
\usepackage{algorithmic}
\usepackage{amsmath}
\usepackage{graphicx}
\usepackage{listings}
\usepackage{xcolor} 
\usepackage{tcolorbox} 
\usepackage{setspace} 
\usepackage{makecell}
\tcbuselibrary{breakable}

\newcommand{\ourbenchmark}{LifeBench}
\newcommand{\cmark}{\ding{51}} 
\newcommand{\xmark}{\ding{55}} 
\renewcommand\footnotetextcopyrightpermission[1]{}
\AtBeginDocument{%
  }

\begin{document}
\tcbset{
    colframe=black!30, 
    colback=gray!3, 
    boxsep=6pt, 
    top=0pt,bottom=0pt,left=0pt,right=0pt, 
    fontupper=\footnotesize\setstretch{1.1}, 
    parbox=false 
}

\lstset{
  language=Python,    
  basicstyle=\small\ttfamily, 
  breaklines=true,    
  commentstyle=\color{gray},  
  keywordstyle=\color{blue}   
}

\title{LifeBench: A Benchmark for Long-Horizon Multi-Source Memory}

\author{Zihao Cheng\textsuperscript{1,2}, Weixin Wang\textsuperscript{1,2}, Yu Zhao\textsuperscript{3}, Ziyang Ren\textsuperscript{3}, Jiaxuan Chen\textsuperscript{1,2}, \\ Ruiyang Xu\textsuperscript{3},  Shuai Huang\textsuperscript{3}, Yang Chen\textsuperscript{3},  Guowei Li\textsuperscript{3}, Mengshi  Wang\textsuperscript{3}, Yi Xie\textsuperscript{3}, Ren Zhu\textsuperscript{3}, \\  Zeren Jiang\textsuperscript{3}, Keda Lu\textsuperscript{3}, Yihong Li\textsuperscript{3},  Xiaoliang Wang\textsuperscript{1}, Liwei Liu\textsuperscript{3}, Cam-Tu Nguyen\textsuperscript{1,2}}
\affiliation{%
   \institution{$^1$State Key Laboratory for Novel Software Technology, Nanjing University}
    \institution{$^2$School of Artificial Intelligence, Nanjing University}
  \institution{$^3$Huawei Technologies Co., Ltd.}
  \city{Nanjing, Jiangsu}
  \country{China}
}

\email{zihao_cheng@smail.nju.edu.cn}

\renewcommand{\shortauthors}{Zihao Cheng et al.}


\begin{abstract}




Long-term memory is fundamental for personalized agents capable of accumulating knowledge, reasoning over user experiences, and adapting across time. However, existing memory benchmarks primarily target declarative memory, specifically semantic and episodic types, where all information is explicitly presented in dialogues. In contrast, real-world actions are also governed by non-declarative memory, including habitual and procedural types, and need to be inferred from diverse digital traces.

To bridge this gap, we introduce \ourbenchmark{}, which features densely connected, long-horizon event simulation. It pushes AI agents beyond simple recall, requiring the integration of declarative and non-declarative memory reasoning across diverse and temporally extended contexts.
Building such a benchmark presents two key challenges: ensuring data quality and  scalability. We maintain data quality by employing real-world priors, including anonymized social surveys, map APIs, and holiday-integrated calendars, thus enforcing fidelity, diversity and behavioral rationality within the dataset. Towards scalability, we draw inspiration from cognitive science and structure events according to their partonomic hierarchy; enabling efficient parallel generation while maintaining global coherence. Performance results show that top-tier, state-of-the-art memory systems reach just 55.2\% accuracy, highlighting the inherent difficulty of long-horizon retrieval and multi-source integration within our proposed benchmark. The dataset and data synthesis code are available at https://github.com/1754955896/LifeBench.

\end{abstract}




\maketitle

\section{Introduction}

Psychology and cognitive science \cite{squire2004memory} have long suggested that memory is not a single entity but a collection of distinct systems that process different kinds of information and operate under different principles. Declarative memory—comprising semantic memory for factual knowledge and episodic memory for personal experiences—supports the ability to recall events and contextual details. In contrast, non-declarative memory facilitates the gradual establishment of skills, habitual behaviors, preference and emotion-conditioned actions. Together, these systems enable humans to both recall the past and adapt behavior based on accumulated patterns, jointly driving daily decision-making and activities.

\begin{figure}[t]
\centering
\includegraphics[width=0.9\linewidth,keepaspectratio]{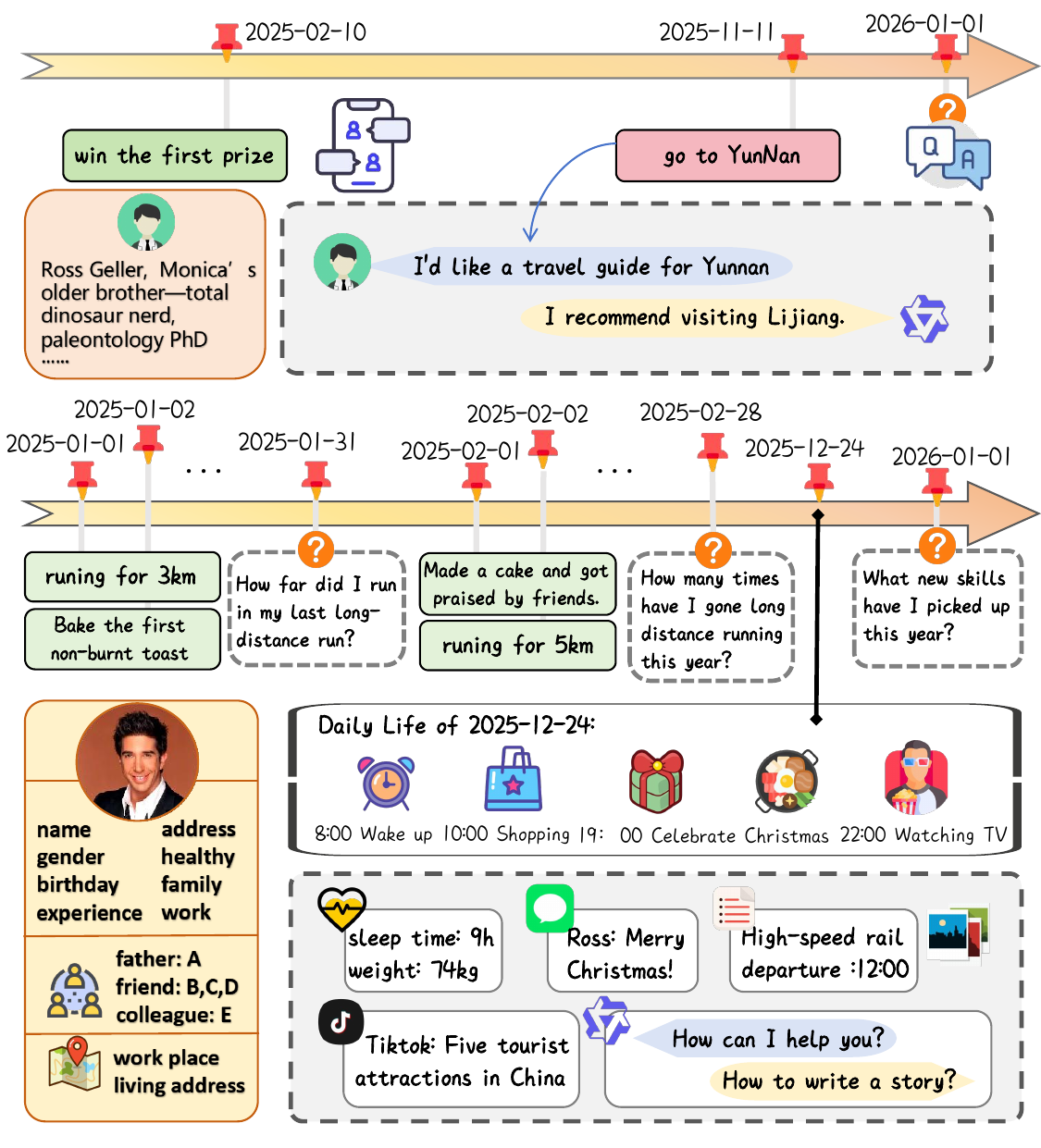}
\caption{\ourbenchmark{} captures rich human activities, which are driven by not only declarative memory (semantic or episodic) but also by non-declarative one (habits, skills, etc.). The user actions and habits need to be inferred from fragmented and diverse digital traces (lower part).}
\label{fig:motivation}
\end{figure}

Recent studies in AI agent have primarily emphasized declarative memory, focusing on retrieving and reasoning over explicitly stored facts or episodic records. This paradigm is exemplified by benchmarks such as LOCOMO \cite{maharana2024locomo}, and LongMemEval \cite{wu2025longmemeval}—and memory-augmented systems, such as HippoRAG \cite{jimenez2024hipporag}, MemOS \cite{li2025memos}.

\begin{table*}[]
    \centering
    \caption{\ourbenchmark{} vs existing memory benchmarks: Ctx depth is defined as the average number of tokens in user history.  Information with (*) is estimated and reported in  \cite{wu2025longmemeval}. Core memory abilities include Information Extraction (IE), Multi-hop reasoning (MR), Temporal and Knowledge Updating (TKU), Nondeclarative Memory Reasoning (ND), and Unanswerable (UA).}
    \begin{tabular}{
    l|c|p{2cm}|c|c|p{1.5cm}|c|c|c|c|c
    }
    \toprule
    \multirow{2}{*}{\textbf{Benchmark}} & \multirow{2}{*}{\textbf{Domain}} & \multirow{2}{*}{\textbf{Artifacts}} & \multirow{2}{*}{\textbf{Events}} &
    \multirow{2}{*}{\textbf{\#Questions}} &
    \multirow{2}{*}{\textbf{Ctx Depth}}  & \multicolumn{5}{c}{\textbf{Core Memory Abilities}}  \\ 
    &  &  &  &  & & 
    \textbf{IE} & \textbf{MR} & \textbf{TKU} & \textbf{ND} & \textbf{UA} \\
    \midrule
    MSC \cite{xu2022beyond} & Open-Domain & Chats & - & - & 1K &  \xmark & \xmark & \xmark & \xmark & \xmark \\
    PerLTQA \cite{du2024perltqa} & Personal & Chats & - & 8593 & 1M (*) &   \cmark & \xmark & \xmark & \xmark & \xmark \\
    LOCOMO \cite{maharana2024locomo} & Personal & Chats & - & 7512 & 10K & \cmark & \cmark & \xmark & \xmark & \cmark \\
    LongMemEval \cite{wu2025longmemeval} & Chats & Chats & - & 500 & 115K, 1.5M &   \cmark & \cmark & \cmark & \xmark & \cmark \\
    Mem-Pal \cite{huang2025mempal} & Personal & Chats, Logs & Sparse & - & - &  \cmark & \cmark & \xmark & \xmark & \xmark \\
    \ourbenchmark{} (ours) & Personal &  Chats, Apps, Health Records & \makecell[c]{Dense}  & 2003 & 3.66M &  \cmark & \cmark & \cmark & \cmark & \cmark \\
    \bottomrule
\end{tabular}
    \label{tab:dataset_comparison}
\end{table*}

While these benchmarks have enhanced the capabilities of memory agents, they offer limited mechanisms for evaluating non-declarative reasoning, which is vital for personalized intelligence. Such knowledge enables systems to align task scheduling with user habits, suggest behavioral adjustments for healthier lifestyles, or optimize device operations for energy efficiency. Furthermore, at scale, anonymized commuting patterns can provide critical insights for urban design. Developing systems that reason over both memory types is therefore essential for building long-term AI agents that understand not just what users know, but the fundamental rhythms of how they live.

Capturing human memory is challenging, as it requires dense, long-term behavioral data that reflect skill acquisition, evolving routines, and causal events. Real-world datasets are often limited by privacy, ethical constraints, and sparsity. To address this, recent work uses synthetic data generation with statistical models \cite{zhang2017privbayes}, GANs \cite{xiong2023trajsgan}, and VAEs \cite{wan2017variational}, and Large Language Models (LLM) \cite{zhang2024guided, ju2025trajllm}. However, these approaches usually focus on isolated behavioral dimensions (e.g. mobility), without capturing the full spectrum of experiences driven by multi-source memory.

This study addresses these limitations by introducing a simulation environment that generates long-horizon, densely connected user experiences governed by human-inspired memory systems. Unlike existing benchmarks that rely solely on explicit dialogue, our framework reflects real-world conditions where memory must be inferred from fragmented  signals such as chats, calendar entries, notes, SMS, and health records (Figure \ref{fig:motivation}). Consequently, \ourbenchmark{} provides a vital resource for developing and evaluating memory agents with tasks that mirror real-world complexity while eliminating the privacy concerns associated with real-user data.

Building such a benchmark presents two fundamental challenges: ensuring data quality and maintaining scalability. To ensure data quality, we incorporate real-world priors such as anonymized social surveys, map APIs, and holiday-integrated calendars. These priors ground the synthesized activities in behavioral rationality and diversity, ensuring that simulated events remain consistent and plausible. To achieve scalability, we organize events within a partonomic hierarchy that enables parallelized simulation while maintaining global coherence. 
This parallelization strategy significantly accelerates the pipeline, reducing the total simulation time from 58 hours to 8 hours per user-year. Performance results show that state-of-the-art systems achieve only 55.2\% accuracy, highlighting the inherent challenge of \ourbenchmark{}.

Our work makes four key contributions:
(1) \textit{\textbf{Human Cognition-Inspired Synthesis}}: we introduce a long-horizon synthesis system that generates year-scale behavioral trajectories grounded in cognitive science. Specifically, the framework models multiple human memory systems and organizes events according to a partonomic hierarchy to maintain consistency across various temporal scales; 
(2) \textit{\textbf{Dataset and Benchmark}}: we construct a multi-source dataset spanning a full year, encompassing personas, daily events, phone artifacts, health records, and monthly summaries to support the modeling of human-like memory. To systematically evaluate reasoning capabilities, we design 2,003 questions across four main categories including information extraction, multi-source reasoning, temporal evolution, non-declarative memory reasoning;
(3) \textit{\textbf{Empirical Evaluation}}: we perform an evaluation of state-of-the-art memory systems such as MemU~\cite{memu}, Hindsight~\cite{latimer2025hindsight}, and MemOS~\cite{li2025memos}. By analyzing their performance and error patterns on our benchmark, we uncover key challenges inherent in long-term memory reasoning and multi-source integration;
(4) \textit{\textbf{Reproducibility Package}}: we release the complete dataset, synthesis framework, and documentation under the Apache License 2.0 to facilitate transparent and responsible research. We invite the community to utilize these toolkits for large-scale data generation and further investigation into personalized intelligence.

\section{Related Work}
\label{sec:related}

Recent years have witnessed growing interest in modeling long-term user history for personalized conversational agents. The MSC dataset \cite{xu2022beyond}  focuses on open-domain chats across multiple sessions. PerLTQA~\cite{du2024perltqa} augments historical chats with social relationships, whereas Mem-Pal~\cite{huang2025mempal} includes both historical dialogues and application logs. LOCOMO~\cite{maharana2024locomo} and LongMemEval~\cite{wu2025longmemeval} benchmark a range of memory reasoning tasks across multiple chat sessions. 
These datasets have laid the groundwork for a variety of memory systems, including A-Mem~\cite{xu2025a-mem}, Zep~\cite{rasmussen2025zep}, Mem0~\cite{chhikara2025mem0}, MemInsights~\cite{salama2025meminsight}, HippoRAG~\cite{jimenez2024hipporag}, MemoryBank~\cite{zhong2024memorybank}, MIRIX~\cite{wang2025mirix}, and Hindsight~\cite{latimer2025hindsight}.

\begin{figure*}[t]
\centering
\includegraphics[width=\textwidth,keepaspectratio]{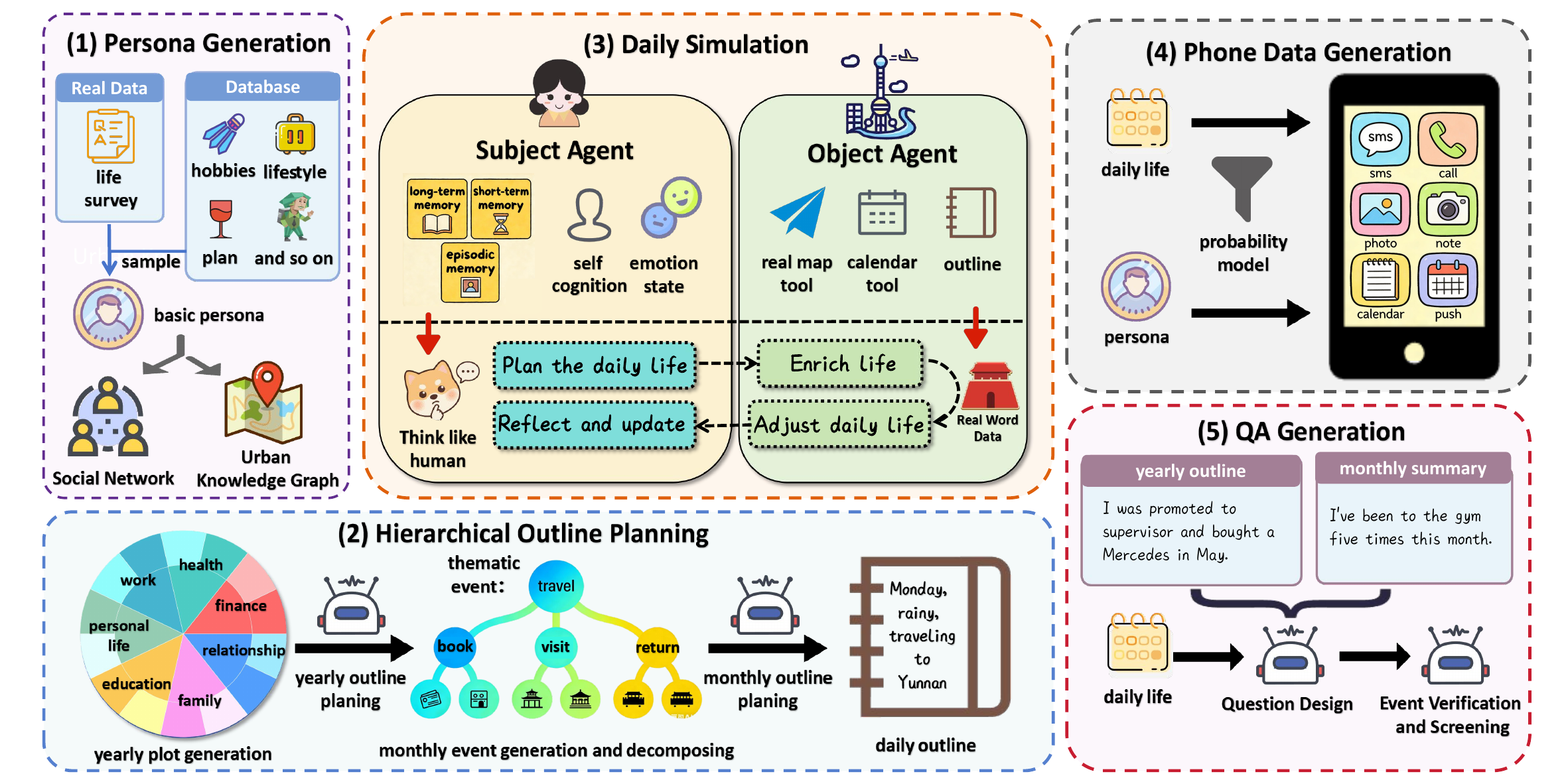}
\caption{The \ourbenchmark{} data synthesis pipeline architecture and key algorithmic components.}
\label{fig:synthesis_pipeline}
\end{figure*}

These benchmarks, however, do not model densely connected events and activities, focusing mainly on semantic and episodic memory, and often assume that user information is explicitly and fully expressed through chats~\cite{xu2022beyond, maharana2024locomo, du2024perltqa, wu2025longmemeval}. In reality, human actions are governed by multiple memory systems and are observable only in fragmented form, distributed across various phone applications. Although Mem-Pal \cite{huang2025mempal} incorporates log data, the application logs are simplified into plain text with explicit event descriptions. Moreover, user events in Mem-Pal are relatively sparse (around 2.5 logs or chats per day). As a result, Mem-Pal fails to capture the real challenge of extracting core memories from  irrelevant data.

A detailed comparison between \ourbenchmark{} and related datasets is provided in Table~\ref{tab:dataset_comparison}. Our benchmark is uniquely characterized by dense event information occurring at a rate of around 14 events per day. These events are partially observable across diverse digital artifacts (24 artifacts per day), which results in a significantly greater context depth exceeding twice the length of the trajectories in LongMemEval. Furthermore, \ourbenchmark{} is the only framework that includes a specific task for non-declarative memory reasoning.


\section{\ourbenchmark{}: Memory Simulation Framework}
\label{sec:dataset}

\subsection{Design Principles}

\textbf{Human Cognition Inspired Synthesis}: our framework is built on two principles: First, we explicitly model how multiple human memory systems \cite{squire2004memory}, including declarative memory (semantic and episodic) and non-declarative memory (habits, skills, emotional conditioning) influence human daily activities. Second, we represent events through a partonomic hierarchy \cite{zacks2001event} that captures their nested structure and ensures consistency across temporal scales.

\textbf{Rationality and Diversity}: We integrate diverse persona-conditioned generation and incorporate multi-source real-world priors, such as maps and calendars, to ensure behavioral diversity and contextual realism in the synthesized data.

\textbf{Phone Artifacts}: Human events and activities can only be observed through their digital traces, such as phone artifacts including calls, messages, calendar entries, agent chats, photos, notes, push notifications, and health records. This highlights a practical challenge that memory agents must address.

\subsection{Synthesis Pipeline Architecture}
Figure~\ref{fig:synthesis_pipeline} illustrates our overall framework for simulating long-horizon, densely connected user activities. The framework comprises five LLM-based modules operating in a pipeline: the \textbf{persona synthesis module} (Section~\ref{subsec:persona}), the \textbf{hierarchical outline planning module} (Section~\ref{subsec:hierarchy}), the \textbf{daily activity simulation module} (Section~\ref{subsec:dual_agent}), and the \textbf{phone data generation module} (Section~\ref{sec:phone-data}), and \textbf{QA generation} (Section~\ref{sec:question-generation}). We use DeepSeek-R1\footnote{https://www.deepseek.com/} as the base LLM in our synthesis framework.

\subsubsection{Persona Synthesis}
\label{subsec:persona}
Each user is modeled with a rich persona capturing demographics (e.g., age, gender, occupation, education, income), lifestyle (e.g., hobbies, habits), personality (e.g., traits, values, emotion), social relationships (e.g., family, friends, colleagues, and interaction patterns), and long-term goals (e.g., career, health, milestones). These personas a generated in three-steps: basic information generation, social networks construction, and real urban location assignments. The specific attribute fields and structures can be found in Appendix ~\ref{app:schema}.

\paragraph{Basic information generation.} 
Base personas are constructed by sampling attributes from various  sources, which are then refined and expanded through LLM rewriting and generation. First, background and lifestyles are sampled from anonymized survey data\footnote{https://www.cnsda.org/index.php}
, which include attributes such as ethnic background, marital status, occupation, and health conditions. Second, additional attributes such as habits, plans, characters,  and skills are sampled from the PersonaHub~\cite{ge2024scaling} dataset. Third, LLMs generates sensitive attributes (fiction names, addresses, etc.), and adjusts sampled fields to maintain consistency. This process grounds each persona in realistic demographic and behavioral priors while ensuring that none of the generated personas correspond to real individuals.


\paragraph{Social network.} Leveraging base personas, LLMs infer and generate social circles for each individual, and define the specific members and their relational networks within each circle. The resulting overall social network comprises approximately 20–30 associated individuals, each equipped with a detailed role profile.

\paragraph{Personalized urban landscape.}
We further construct a personalized urban landscape  for each individual to ground their activities in real-world geography. Using the map API\footnote{https://lbs.amap.com/} and LLMs, we map each persona's geographical information real-world locations within their respective city. Based on these mappings, we identify plausible anchor locations that serve as the primary sites of each individual’s daily activities.

\subsubsection{Hierarchical Outline Planning}
\label{subsec:hierarchy}

We construct a personal event partonomic tree to support the modeling of user activities. Higher-level nodes represent long-span thematic events, capturing a participant’s intentions and plans, whereas leaf nodes represent atomic events, defined as those shorter than one day and closely related to daily activities. We first describe the procedure for generating the event tree, and then explain how daily activities are subsequently derived from it (Section~\ref{subsec:dual_agent}).

\paragraph{Plot Generation.} To ensure coherence, we first establish a plot outline, a global story that serves as the narrative of an individual. Using a predefined database of approximately 50 candidate plots of seven categories (work, health, finance, relationship, family, education, and personal life), the LLM selects up to 10 plots tailored  to the user’s persona. To allow for controlled synthesis, our simulation framework supports customization via priority labels, enabling human-assigned priors to weight the selection of specific plots. The selected plots are ultimately integrated through temporal sequencing and optimization to produce a seamless, one-year chronological outline. The plot schema can be found in Appendix  ~\ref{app:schema}.



\paragraph{Thematic Event Generation and Decomposition.} Guided by the plot outline, the LLM generates thematic events month-by-month in an autoregressive manner. Each event includes attributes such as location, participants, duration, start and end dates. We assign thematic events to a month if their start dates fall within the given month window, though events may naturally span across multiple months. To ensure narrative and temporal consistency, the prompt includes the global plot outline, a summary of the preceding month’s events, and relevant holiday data. Finally, we directly ask the LLM to decompose each thematic event into nested subevents until reaching atomic events, which are defined as those lasting less than one day, thereby constructing the event partonomic tree.



 

\paragraph{Atomic Event Optimization.} 
Atomic events generated in the preceding phase are all mapped to their corresponding thematic events. Nevertheless, certain atomic events are inherently short-term in nature, and all events are susceptible to contextual factors including emotional states, self-cognition such as hobbies, and habits (\textit{\textbf{non-declarative memory}}). To capture these dynamics and refine atomic events, the LLM analyzes and adjusts those within each two-week time window based on contextual factors derived from the user’s persona and history. New atomic events can then be introduced to reflect users’ daily routines such as regular physical exercise, while existing ones can be modified to better align with contextual conditions like weather conditions and personal moods.

\subsubsection{Dual-Agent Daily Activity Simulation}
\label{subsec:dual_agent}

Daily activities can be viewed as the smallest units of events, typically lacking explicit goals or plans. Each atomic event may consist of one or more daily activities, and each activity may or may not has digital traces on the user phone (referred to as phone artifacts). To generate these daily activities, we adopt a dual-agent simulation framework: a subjective agent simulates human-like reasoning to produce plausible activities, while an objective agent verifies and grounds these activities in real-world data to ensure fidelity and consistency.

\paragraph{Subjective Agent:}
The Subjective Agent is an LLM that simulates human behavior to generate daily activities conditioned on atomic events, user profiles, and daily contextual factors  (\textit{\textbf{non-declarative memory}}). The LLM is equipped with three types of memory: \textit{long-term memory}, derived from summaries of the preceding month and the current month; \textit{short-term memory}, which stores detailed records of activities from the last seven days; and \textit{episodic memory}, comprising the three most relevant daily activities retrieved from the accumulated storage of past activities (here, the day's atomic events are used as the query for retrieval).

\paragraph{Objective Agent:} 
The Objective Agent enforces physical and temporal constraints, validating the feasibility of daily plans by checking for time conflicts, travel times via map APIs, venue availability, and consistency with established facts, while also introducing supplementary events and external environmental feedback. By doing so, the Objective Agent ensures the rationality and authenticity of the simulated activity data. 

\paragraph{Overall Procedure:} \textit{The overall daily simulation follows a day-by-day autoregressive manner, whereas the simulation for each day proceeds iteratively as follows}: the Subjective Agent first proposes a daily plan based on scheduled atomic events, emotional state, habitual cognition; the Objective Agent then validates the plan, provides feedback, generates sparse time slots for supplementary events, and assigns real location and date information; finally, the Subjective Agent updates and optimizes the plan according to this feedback, refining the associated attribute information.

\subsubsection{Phone data generation.} 
\label{sec:phone-data}

Phone data artifacts serve as the raw observations for inferring underlying user activities, events, and habits. The main types of phone data are as follows:

\begin{itemize}
\item Contacts: Generated based on the persona’s social network.
\item SMS, Calls, Push Notifications, Photos, Notes, Calendar, Agent-chats: The LLM analyzes daily activities and estimates the likelihood of each type of mobile operation. Sampling is then applied based on the estimated likelihood, and the LLM generates the corresponding data. To simulate realistic interpersonal conversations, the LLM also generates topic-specific SMS content unrelated to personal events—such as discussions of third-party incidents, social topics, personal thoughts, or online content.
\item Noise Data: Event-irrelevant mobile content, including random conversations, advertisements, web browsing records, and device notifications—some meaningless, others containing user information—is synthesized to simulate realistic background activity.
\end{itemize}
The complete phone data schema is provided in Appendix ~\ref{app:schema}. Due to the presence of noise and the fact that not all daily activities are captured in phone data, this introduces a unique challenge for benchmarking  memory systems.

\paragraph{Health and Sports Records} Daily health data include step counts, travel distances, and workout records (gym, running, cycling, swimming) with associated duration, intensity, and calorie consumption; sleep duration annotated with categorical tags (restful, interrupted); daily activity summaries; and physical recovery indicators. All health metrics are generated conditioned on daily activities, the daily contextual factors (persona’s habits, and fluctuations in daily emotion and stress levels). The full Health and Sports Records data schema is provided in Appendix \ref{app:schema}.

\subsubsection{Question Answering Generation} \label{sec:question-generation} An in-progress question generation mechanism produces queries dynamically during daily activity simulation rather than post hoc, enabling exact grounding and evidence assignment from corresponding phone data. The question design aims to address several distinct challenges that have been largely overlooked in existing memory benchmarks. Specifically, it targets temporal evolution testing, where time-dependent queries assess whether models can integrate new information as data accumulates; causal reasoning, which requires linking temporally distant events to infer underlying causes; and ambiguous concept handling, which tests the ability to manage updates and distinctions in attributes, names, and semantic nuances. We further include nondeclarative memory reasoning questions that probe emotions, habits, and preferences, requiring models to infer patterns from long-range daily activities.

\begin{figure}[t]
\centering
\includegraphics[width=0.47\textwidth,keepaspectratio]{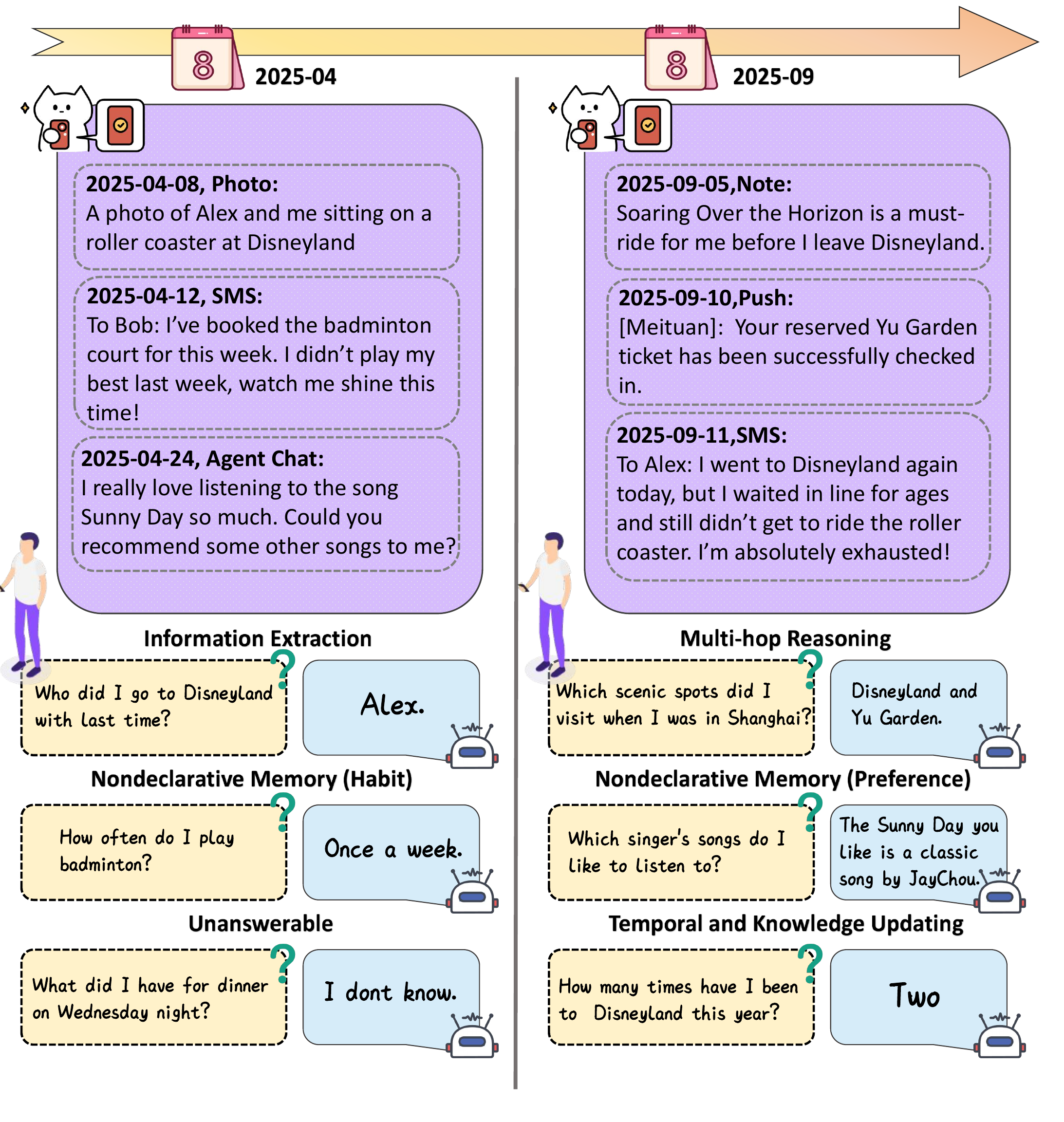}
\caption{Samples of \ourbenchmark{} question types. }
\label{fig:question_samples}
\end{figure}

Our benchmark includes five question categories with sample questions are shown in Figure \ref{fig:question_samples}. Each question category is designed to evaluate a distinct aspect of memory:

\begin{itemize}
\item Information Extraction (IE): Tests direct fact retrieval from individual memory records, such as locating the time of a specific gym session.
\item Multi-hop Reasoning (MR): Requires aggregating and reasoning over multiple memory points, for example counting how many times the user met colleagues from the marketing team in Q1.
\item Temporal and Knowledge Updating (TKU): Assesses temporal reasoning and the ability to handle evolving information, such as updating the total number of swimming sessions as time progresses.
\item Nondeclarative Memory Reasoning (ND): Evaluates system understanding of user habits, skills, emotion and preferences, e.g., identifying the user’s usual gym routine.
\item Unanswerable (UA): Contains questions that cannot be inferred from available data or are factually false, requiring the model to respond that the query is unanswerable (e.g., “How many bottles of milk did I drink on May 8th?”).
\end{itemize}

Each category includes two question formats. Open-ended QA prompts open-ended responses to evaluate completeness, coherence, and the appropriate level of abstraction. Multiple-choice QA provides four candidate answers plus a “not in memory” option to test discriminative ability and reduce hallucination. Paired questions targeting the same memory content allow direct comparison between recall and recognition performance.

\paragraph{Phone Data Supplementation:} After formulating the questions, we analyze the related phone data to assess whether it provides sufficient information to answer the questions. If gaps are found, additional phone data are generated to fill in the missing details, ensuring that all questions—except unanswerable ones—can be answered based on the phone data.

\paragraph{Data Filtering and Manual Checking:} An initial screening is performed using an LLM to remove questions that are unreasonable or whose answers can be directly inferred from the question itself. The remaining questions are then manually reviewed, with unreasonable items removed and answers and options rewritten or optimized as needed. This process produces the final QA dataset.

\subsection{Scalability}
\label{subsec:Scale}
A major challenge in LLM-based simulations of long-horizon, causally linked events is the high computational cost. A naive implementation based on sequential generation takes 58 hours to simulate one year of activities for a single user, with 3-4 hours for hierarchical outline planing, 40 hours for daily activity generation, and 14 hours for phone data generation. To overcome this, we accelerate the simulation based on the following key insights.

\paragraph{Parallel thematic event decomposition} In the previous section, thematic events are decomposed independently, but the degree of parallelism is limited by the number of thematic events. To accelerate this process, we design a parallelization method that maintains an event tree and recursively decomposes each “splittable” leaf node into subevents. This approach increases the degree of parallelism to the number of leaf nodes in the tree. A detailed description is provided in Algorithm \ref{alg:core_event_decomposition}  in Appendix ~\ref{app:scale}. We ultimately achieved an acceleration of event decomposition from approximately 2 hours to around 30 minutes through 24-thread parallelization.


\paragraph{Time-sliced daily simulation} The primary bottleneck lies in simulating daily activities. Using a day-by-day autoregressive approach, generating data for a single individual takes approximately 40 hours. To accelerate this process, we assume that activities separated by two weeks can be treated independently when conditioned on their corresponding atomic events. This assumption allows activities from different two-week periods to be generated in parallel, while those within the same period remain sequential. Since explicit activity memory is limited in this parallel setting, we supplement the long-term, short-term, and episodic memories with the corresponding atomic events, which were previously generated during the event decomposition stage. We ultimately completed the processing of one-year daily simulation in approximately 2 hours via 24-thread parallelization.


\paragraph{Parallel phone and health data generation} Mobile phone data and health records depend solely on daily activities and contextual factors, which were previously generated during the atomic event generation and daily activity simulation steps; therefore, data for each day can be generated independently in parallel. This process takes approximately 1-2 hours, keeping the total simulation time for one person under 8 hours.

\subsection{Data Statistics and Quality Assessment}
\subsubsection{Data Statistics} 
\begin{figure}[t]
\centering
\includegraphics[width=\linewidth,keepaspectratio]{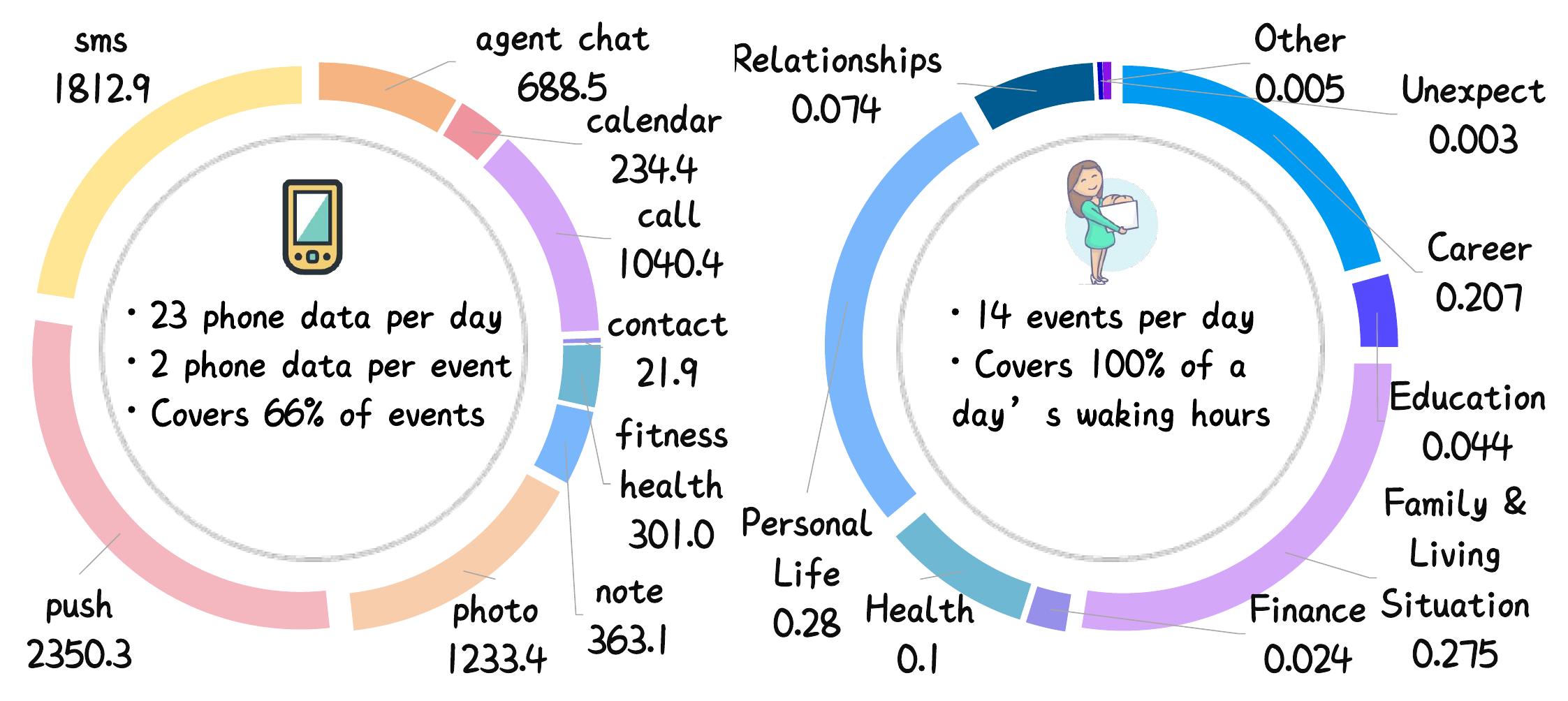}
\caption{Statistics of Events and Phone Data.}
\label{fig:brief-statistics}
\end{figure}

Using the data simulation framework, we synthesize one year long of user activities for 10 users. Each user has a total of 5,149 events, averaging roughly 14 events per day, with a diverse distribution illustrated in Figure~\ref{fig:brief-statistics}. Additionally, each user has 8046 phone data and health records, with their distribution shown on the left of Figure~\ref{fig:brief-statistics}. Full data statistics are provided in Table~\ref{tab:dataset_stats} in Appendix.

\begin{figure}[t]
\centering
\includegraphics[width=0.85\linewidth,height=2.7cm]{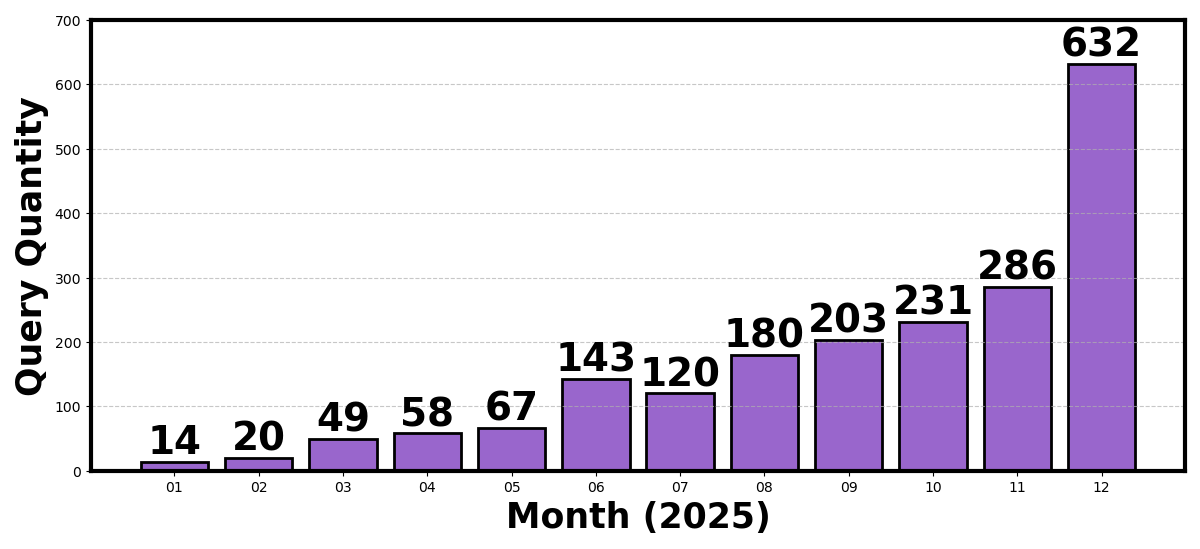}
\caption{Distribution of Query Time.}
\label{fig:query_time_evaluation}
\end{figure}

\paragraph{Question Statistics} A total of $2,003$ questions cover five categories, with the distribution shown in Table \ref{tab:query_information}. Each question is associated with a timestamp to evaluate the system’s ability to capture memory evolution, and the temporal distribution of the questions is illustrated in Figure \ref{fig:query_time_evaluation}.

\begin{table}[t]
    \centering
    \caption{Statistics on QA pairs of \ourbenchmark{}}
    \label{tab:query_information}
    \begin{tabular}{@{}lcc@{}}
        \toprule
        \textbf{Type} & \textbf{Subtype} & \textbf{Ques. (Percent)} \\
        \midrule
        \multirow{5}{*}{\textbf{Category}} & IE & 718 (35.85\%) \\
        & MR & 597 (29.81\%) \\
        & TKU & 229 (11.43\%) \\
        & ND & 429 (21.42\%) \\
        & UA & 30 (1.50\%) \\
        \midrule
        \multirow{2}{*}{\textbf{Format}} & Multiple Choice & 517 (25.81\%) \\
        & Short Answer & 1486 (74.19\%) \\
        \midrule
        \textbf{Total} & \textbf{} & \textbf{2003 (100.00\%)} \\
        \bottomrule
    \end{tabular}
\end{table}

\subsubsection{Data Quality Evaluation} We employ automated and manual evaluations to assess the consistency, authenticity, and diversity of the synthesized event data. 

\paragraph{Manual Evaluation.} We evaluated the \textit{rationality} of daily activity and phone data for five individuals over one month (about 400–500 activities per user). For daily activities, we checked for temporal coherence, content, location, and participant consistency, marking any unreasonable entries as invalid. For phone data, each data entry was matched to the corresponding activities and assessed for consistency and rationality, with invalid entries recorded. The proportions of valid data entries were then calculated.

For diversity evaluation, we exploited a 10-point scale based on five criteria: (1) inter-individual variability, (2) intra-individual variability across days, (3) non-stereotypical representativeness: a balanced mix of individual-specific and common events that avoid rigid character stereotypes, (4) fine-grained metric diversity: fluctuation in fine-grained metrics such as sleep or diet, and (5) plausibility: reflecting the complexity and unpredictability of real life while remaining coherent. Each criterion contributes up to 2 points, totaling 10 points. The detailed information for manual evaluation rubrics is given in Appendix ~\ref{app:manual evaluation}.

\paragraph{Automatic Evaluation.} We used LLM and the AMap API to automatically evaluate the generated event data along four dimensions: \textit{relation consistency} (alignment with predefined social relations), \textit{location authenticity} (verification of places against real-world maps), \textit{trip authenticity} (plausibility of travel durations and routes), and \textit{event diversity} (balance and variety of event types measured by Normalized Shannon Entropy and Simpson Diversity). Higher scores across all dimensions indicate greater realism and diversity in the generated data. Detailed definitions and computation procedures are provided in Appendix~\ref{app:evaluation}.




\begin{figure}[t]
\centering
\includegraphics[width=0.45\textwidth,keepaspectratio]{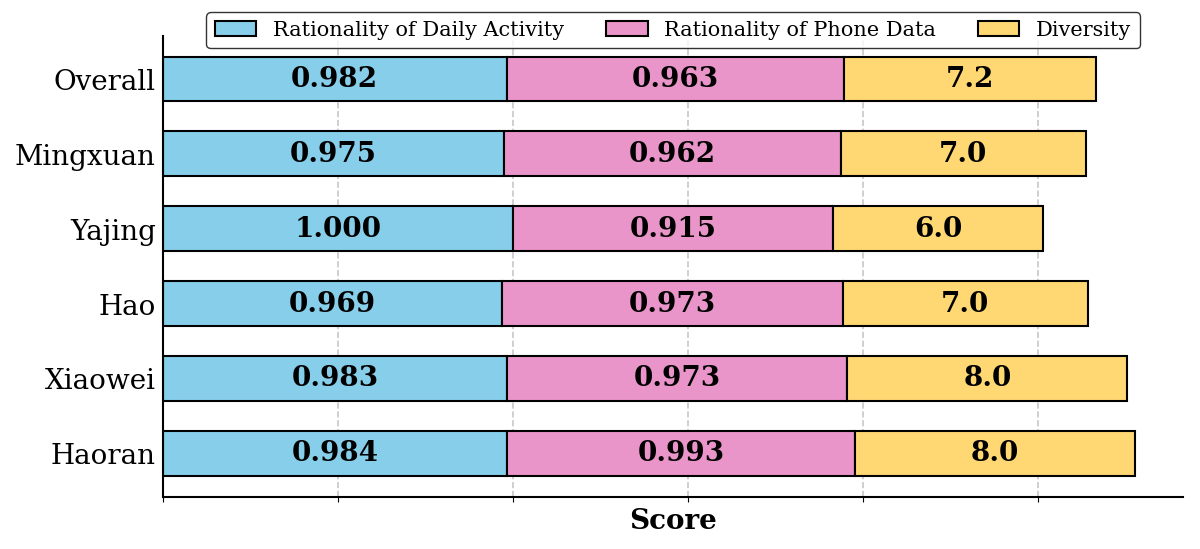}
\caption{Manual Evaluation on Synthetic Data.}
\label{fig:manual-evaluation}
\end{figure}



\subsubsection{Quality Assessment Results} 

\begin{table}[t]
\centering
\caption{Automatic Data Quality Assessment Results}
\label{tab:quality_stats}
\renewcommand{\arraystretch}{1.0}
\begin{tabular}{@{}p{0.2\linewidth}p{0.5\linewidth}r@{}}
\toprule
\textbf{Category} & \textbf{Metric} & \textbf{Value} \\
\midrule
\textit{Consistency} & Relation consistency & 0.83 \\
\midrule
\multirow{2}{*}{\textit{Authenticity}} & Location authenticity & 0.83 \\
                                      & Trip authenticity & 0.75 \\
\midrule
\multirow{2}{*}{\textit{Diversity}} & Normalized Shannon Entropy & 0.75 \\
                                    & Simpson Diversity Index & 0.78 \\
\bottomrule
\end{tabular}
\end{table}

Figure~\ref{fig:manual-evaluation} presents the manual evaluation results. Overall, the synthesized data demonstrates high rationality (0.962–1.0) across both daily activity and phone data. Minor irrationalities mainly stem from repeated events or step jumps due to hallucinations. No major logical inconsistencies are observed, and most content remains coherent and authentic. In terms of diversity, the data performs well in both inter- and intra-individual variability but shows room for improvement in fine-grained metric diversity and the non-stereotypical representativeness.

Table~\ref{tab:quality_stats} presents the automated assessment of synthetic data quality. Relational consistency and location authenticity meet the expected levels, reflecting the effectiveness of our design that integrates human-curated personas with real-world priors. Trip authenticity is slightly lower, likely due to temporal constraints in the real-world datasets, which limit access to transportation conditions for specific simulated periods. Regarding diversity, both Normalized Shannon Entropy (0.75) and Simpson Diversity Index (0.78) show moderately high diversity, i.e. the generated events are diverse with only mild concentration in certain categories. 

\section{Evaluating Memory Systems on \ourbenchmark{}}
\label{sec:systems}

\subsection{Memory Systems}
To illustrate the difficulty of \ourbenchmark{}, we conduct empirical evaluations with a number of state-of-the-art memory systems that have recently achieved top performance on existing benchmarks, including LoCoMo~\cite{maharana2024locomo} and LongMemEval~\cite{wu2025longmemeval}. These systems represent the current frontier of long-term memory modeling in LLMs, covering diverse architectures and retrieval mechanisms. By testing them under our \ourbenchmark{}, we aim to examine their drawbacks in more complex, temporally extended scenarios.

\textbf{MemU}~\cite{maharana2024locomo} is a memory framework for 24/7 proactive agents, using a dual-agent setup where a primary agent executes tasks while a memory agent (MemU Bot) monitors interactions, summarizes memories, and predicts user intentions. 


\textbf{Hindsight}~\cite{latimer2025hindsight} structures memory agent into four networks: world facts, personal experiences, synthesized observations, and evolving opinions. This design separates raw data from inferred knowledge. It implements three core operations (retain, recall, reflect) through TEMPR (Temporal Entity Memory Priming Retrieval) for \textit{memory storage and retrieval}, and CARA (Coherent Adaptive Reasoning Agents) for \textit{preference-conditioned reflection}. The system uses multi-strategy retrieval combining semantic, graph, temporal, and maintains an opinion network with updated confidence scores.


\textbf{MemOS}~\cite{li2025memos} serves as a memory operating system that \textit{integrates multi-source LLM memories} such as plaintext, activation (KV-cache), and LLM parametric memories within a unified “MemCube” abstraction. It manages the entire memory lifecycle with metadata recording source history, version control, and access policies. It adopts a three-tier architecture comprising working memory, long-term storage, and a cold archive, with dynamic memory migration across tiers. An intent-aware scheduler proactively preloads or fuses relevant memories based on anticipated task demands.


\subsection{Experimental Setup}
We evaluate the memory systems using a unified base LLM, standardized input formats, and a consistent evaluation protocol.

\paragraph{Base LLM and Data Processing}
GPT-5.1-Mini \cite{openai_gpt51mini} is used for both memory operations and answer generation, while text-embedding-3-small \cite{openai_textembedding} handles text embeddings. As for input data, we convert our dataset to the standard format of LoCoMo test configurations~\cite{maharana2024locomo, latimer2025hindsight, li2025memos}, which are supported by all the memory systems. Since existing benchmarks are primarily designed for conversational text and cannot directly process structured mobile phone data (e.g., contacts, calls, SMS, calendar entries, agent chats, photos, notes, push notifications, and health data), we create a summarized info field for structured types. This field condenses contextual information into textual form, enabling easier processing by current memory systems. Future memory systems may choose to leverage either the summarized or raw structured data.

\paragraph{Evaluation Protocol}
We use an LLM-as-judge approach, employing GPT-5.1-Mini as the judge LLM. The standard LoCoMo evaluation prompt~\cite{maharana2024locomo} is used to determine if a model correctly answers each question. The model score is calculated as the proportion of its correctly answered questions.

\subsection{Experimental Results}
\label{sec:results}

\begin{figure}[t]
\centering
\includegraphics[width=\linewidth, height=3.5cm]{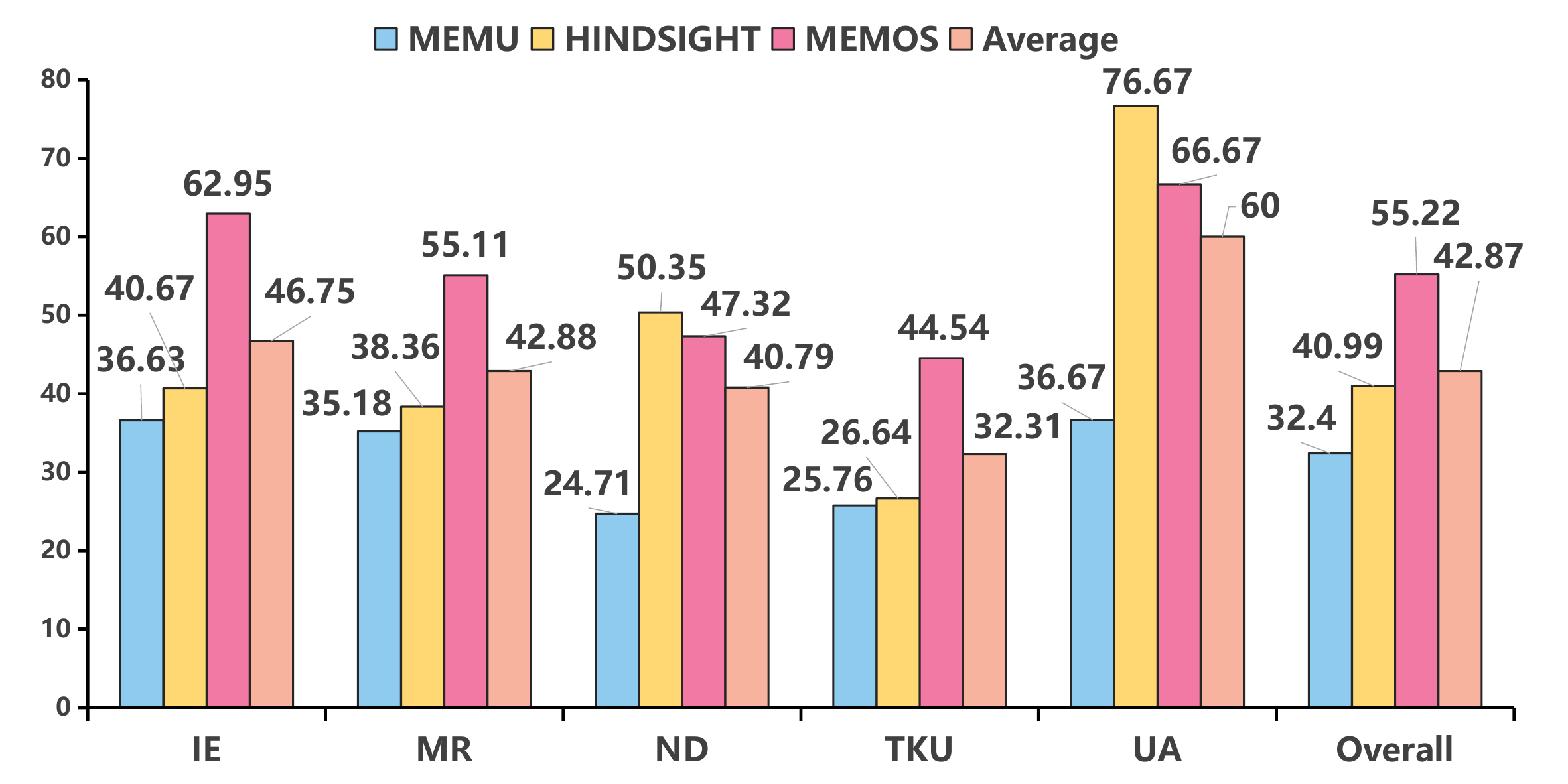}
\caption{Performance of Memory Systems on \ourbenchmark{}.}
\label{fig:main_results}
\end{figure}

Figure~\ref{fig:main_results} shows overall accuracy across all question types as well as by category. The top-performing model, MemOS, achieves only 55.22\% overall accuracy, with Hindsight ranking second at 40.99\%. Notably, Hindsight scores around 90\% on existing memory benchmarks (LoCoMo and LongMemEval) \cite{latimer2025hindsight}, highlighting that current benchmarks are no longer sufficient for testing state-of-the-art memory systems due to their simplified assumptions. In contrast, \ourbenchmark{} presents a much-needed challenge to drive further progress in memory system research.

Although MemOS is the top-performing model on average, its performance varies across QA categories, as shown in Figure~\ref{fig:main_results}. Specifically, MemOS outperforms Hindsight on IE, MR, and TKU, but falls behind on Non-Declarative Memory 
Reasoning (ND) and Unanswerable (UA) questions. Its strong results on factual and temporal questions (IE, MR, TKU) is likely due to the benefit of integrating multiple source LLM memories (plaintext, KV cache, LLM model parameters) within the unified MemCube. In addition, MemOS’s success on TKU questions demonstrates the effectiveness of its memory version management in handling evolving memories.

Despite ranking second overall, Hindsight excels on questions about  preferences (a part of ND type), thanks to its ability to model preference-conditioned reflection and opinion networks. Additionally, its higher performance on UA reflects its tendency to refrain from answering ambiguous queries. Conversely, MemOS attempts to answer most questions, which lowers its UA scores but allows it to perform better on other categories.

Among the three systems, MemU performs significantly worse than Hindsight and MemOS, despite achieving better results over Hindsight on LoCoMo \cite{li2025memos}. This is mainly because MemU often produces brief memory summaries that fail to capture the complexity and nuances of long-horizon daily activities in \ourbenchmark{}.

\subsection{Error Analysis}
This section provides a qualitative analysis to draw insights into the memory reasoning abilities of MemOS and Hindsight.

The first case study (Question 1) is presented in Table~\ref{tab:memos-vs-hindsight}, where MemOS provides the correct answer while Hindsight does not. Analysis of the retrieved memories reveals that Hindsight’s outputs omit the event of making fabric kitten pendants, focusing instead on general content related to cats and neighbors. The retrieval strength of MemOS originates from its design to standardize multi-source (photos, sms) in a standardized memory units (the MemCube). Further examination of their reasoning processes shows that Hindsight overlooked temporal cues such as “right after” and key relational terms like “parent–child,” resulting in an incorrect response based on an unrelated cat event from a different month. MemOS, however, successfully incorporated these contextual and temporal constraints to produce the correct reasoning outcome.

\begin{table}[t]
    \centering
    \caption{Case Studies of Memos vs Hindsight}
    
    \begin{tabular}{p{8cm}}
        \toprule
         \textbf{Question 1}: After the neighbor came to express their gratitude, what cat-themed parent-child activity did I do right after? \\
         \setlength{\parindent}{0.5em} A: Went to the neighbor's house with my daughter... \\
         \setlength{\parindent}{0.5em} B: Told the whole family about the experience ... the cat.\\
         \setlength{\parindent}{0.5em} C: Watched cat videos and made fabric kitten pendants. \\
         \setlength{\parindent}{0.5em} D: Discussed the possibility of keeping a pet in the future. \\
         \setlength{\parindent}{0.5em}\textbf{\textit{MemOS}}: C; \textbf{\textit{Hindsight}}: A,
         \textbf{\textit{Groundtruth}}: C  \vspace{=.5em} \\
         \textbf{Question 2}: How was Ma Xiulan's mood or state when she took care of her neighbor's cat independently for the first time?\\
         \leftskip=0.5em \textbf{\textit{MemOS}}: Happy and pleasantly surprised \\ 
         \leftskip=0.5em \textbf{\textit{Hindsight}}: She was initially a little nervous, scared and cautious, but after finishing the first feeding and litter box cleaning, she felt at ease and built up a sense of security and confidence. Over the following days, the cat gradually relaxed ... \\
        \leftskip=0.5em \textbf{\textit{Groundtruth}}: She felt a little nervous and her movements were unskilled, but she successfully completed the task.\\ 
        \bottomrule
    \end{tabular}
    \label{tab:memos-vs-hindsight}
\end{table}

The second case study (Question 2) is presented in Table~\ref{tab:memos-vs-hindsight}, where Hindsight produces the correct answer while MemOS does not. Analysis of the retrieved results shows that both systems identified the event of taking care of the cat for the first time. However, Hindsight’s memory units include richer descriptions of the user’s psychological states and behaviors, whereas MemOS focuses primarily on factual event records. During response generation, Hindsight tends to analyze the user’s emotions and reasoning to produce more detailed and comprehensive answers, while MemOS favors concise factual statements. More case studies are given in Appendix \ref{errorcase} that reveal common errors of MemOS and Hindsight include failure to retrieve correct evidence, incomplete evidence retrieval, hallucinated content, incorrect reasoning, and memory omission. 

\textbf{Implications for Future Memory System Design}: A system that addresses multiple facets of human memory is essential. Unfortunately, matching, linking, and aligning memories across different data sources and memory types remains a significant challenge. In addition, temporal reasoning and knowledge updating remain major bottlenecks because current systems rely mainly on similarity-based retrieval and lack mechanisms for temporal constraints or reasoning. Developing specialized temporal indexing and time-aware query processing is therefore necessary to accurately track the evolution of memory. Furthermore, a large scale of synthetic data can facilitate memory training \cite{li2025memos}. The implementation of such systems must prioritize ethical and responsible use, as discussed further in Appendix \ref{sec:ethics}.


\section{Conclusion}
\label{sec:conclusion}

In this paper, we introduce \ourbenchmark{}, a synthetic data framework capable of generating long-term, densely connected personal events and activities. Inspired by cognitive science, our framework is built on two principles: 1) modeling multiple human memory systems, including declarative and non-declarative memory, in shaping human actions, and 2) representing events through a partonomic hierarchy that captures their decomposable structure. We also design a parallelization solution that reduces the synthesis time for year-long dense activities for a single individual from 58 hours to 8 hours.

Based on this framework, we develop a year-long dataset covering 10 users, including detailed personas and activities with underlying causal events. Evaluations of state-of-the-art memory systems reveal the challenges posed by \ourbenchmark{}, highlighting its value as a resource for advancing AI memory systems. These benchmarks and datasets can also support future applications in personalized AI assistants, digital health coaching, behavioral research. We release the complete dataset, generation code, and documentation to support reproducible and responsible research.


\bibliographystyle{ACM-Reference-Format}
\bibliography{main}


\begin{thebibliography}{35}


\ifx \showCODEN    \undefined \def \showCODEN     #1{\unskip}     \fi
\ifx \showISBNx    \undefined \def \showISBNx     #1{\unskip}     \fi
\ifx \showISBNxiii \undefined \def \showISBNxiii  #1{\unskip}     \fi
\ifx \showISSN     \undefined \def \showISSN      #1{\unskip}     \fi
\ifx \showLCCN     \undefined \def \showLCCN      #1{\unskip}     \fi
\ifx \shownote     \undefined \def \shownote      #1{#1}          \fi
\ifx \showarticletitle \undefined \def \showarticletitle #1{#1}   \fi
\ifx \showURL      \undefined \def \showURL       {\relax}        \fi
\providecommand\bibfield[2]{#2}
\providecommand\bibinfo[2]{#2}
\providecommand\natexlab[1]{#1}
\providecommand\showeprint[2][]{arXiv:#2}

\bibitem[Bush et~al\mbox{.}(1945)]%
        {bush1945we}
\bibfield{author}{\bibinfo{person}{Vannevar Bush} {et~al\mbox{.}}} \bibinfo{year}{1945}\natexlab{}.
\newblock \showarticletitle{As we may think}.
\newblock \bibinfo{journal}{\emph{The atlantic monthly}} \bibinfo{volume}{176}, \bibinfo{number}{1} (\bibinfo{year}{1945}), \bibinfo{pages}{101--108}.
\newblock


\bibitem[Chhikara et~al\mbox{.}(2025)]%
        {chhikara2025mem0}
\bibfield{author}{\bibinfo{person}{Prateek Chhikara}, \bibinfo{person}{Dev Khant}, \bibinfo{person}{Saket Aryan}, \bibinfo{person}{Taranjeet Singh}, {and} \bibinfo{person}{Deshraj Yadav}.} \bibinfo{year}{2025}\natexlab{}.
\newblock \showarticletitle{Mem0: Building production-ready ai agents with scalable long-term memory}.
\newblock \bibinfo{journal}{\emph{arXiv:2504.19413}} (\bibinfo{year}{2025}).
\newblock


\bibitem[Du et~al\mbox{.}(2024)]%
        {du2024perltqa}
\bibfield{author}{\bibinfo{person}{Yiming Du}, \bibinfo{person}{Hongru Wang}, \bibinfo{person}{Zhengyi Zhao}, \bibinfo{person}{Bin Liang}, \bibinfo{person}{Baojun Wang}, \bibinfo{person}{Wanjun Zhong}, \bibinfo{person}{Zezhong Wang}, {and} \bibinfo{person}{Kam-Fai Wong}.} \bibinfo{year}{2024}\natexlab{}.
\newblock \showarticletitle{{P}er{LTQA}: A Personal Long-Term Memory Dataset for Memory Classification, Retrieval, and Fusion in Question Answering}. In \bibinfo{booktitle}{\emph{Proceedings of the 10th SIGHAN Workshop on Chinese Language Processing (SIGHAN-10)}}. \bibinfo{pages}{152--164}.
\newblock


\bibitem[Ge et~al\mbox{.}(2024)]%
        {ge2024scaling}
\bibfield{author}{\bibinfo{person}{Tao Ge}, \bibinfo{person}{Xin Chan}, \bibinfo{person}{Xiaoyang Wang}, \bibinfo{person}{Dian Yu}, \bibinfo{person}{Haitao Mi}, {and} \bibinfo{person}{Dong Yu}.} \bibinfo{year}{2024}\natexlab{}.
\newblock \showarticletitle{Scaling synthetic data creation with 1,000,000,000 personas}.
\newblock \bibinfo{journal}{\emph{arXiv:2406.20094}} (\bibinfo{year}{2024}).
\newblock


\bibitem[Gemmell et~al\mbox{.}(2006)]%
        {gemmell2006mylifebits}
\bibfield{author}{\bibinfo{person}{Jim Gemmell}, \bibinfo{person}{Gordon Bell}, {and} \bibinfo{person}{Roger Lueder}.} \bibinfo{year}{2006}\natexlab{}.
\newblock \showarticletitle{MyLifeBits: a personal database for everything}.
\newblock \bibinfo{journal}{\emph{Commun. ACM}} \bibinfo{volume}{49}, \bibinfo{number}{1} (\bibinfo{year}{2006}), \bibinfo{pages}{88--95}.
\newblock


\bibitem[Guo et~al\mbox{.}(2018)]%
        {guo2018user}
\bibfield{author}{\bibinfo{person}{Ao Guo}, \bibinfo{person}{Jianhua Ma}, \bibinfo{person}{I Kevin}, {and} \bibinfo{person}{Kai Wang}.} \bibinfo{year}{2018}\natexlab{}.
\newblock \showarticletitle{From user models to the cyber-i model: approaches, progresses and issues}. In \bibinfo{booktitle}{\emph{Proceedings of 16th Intl Conf on Dependable, Autonomic and Secure Computing, 16th Intl Conf on Pervasive Intelligence and Computing, 4th Intl Conf on Big Data Intelligence and Computing and Cyber Science and Technology Congress}}. \bibinfo{pages}{33--40}.
\newblock


\bibitem[Huang et~al\mbox{.}(2026)]%
        {huang2025mempal}
\bibfield{author}{\bibinfo{person}{Zhaopei Huang}, \bibinfo{person}{Qifeng Dai}, \bibinfo{person}{Guozheng Wu}, \bibinfo{person}{Xiaopeng Wu}, \bibinfo{person}{Kehan Chen}, \bibinfo{person}{Chuan Yu}, \bibinfo{person}{Xubin Li}, \bibinfo{person}{Tiezheng Ge}, \bibinfo{person}{Wenxuan Wang}, {and} \bibinfo{person}{Qin Jin}.} \bibinfo{year}{2026}\natexlab{}.
\newblock \showarticletitle{Mem-PAL: Towards Memory-based Personalized Dialogue Assistants for Long-term User-Agent Interaction}. In \bibinfo{booktitle}{\emph{Proceedings of the AAAI Conference on Artificial Intelligence}}.
\newblock


\bibitem[Huang et~al\mbox{.}(2025)]%
        {huang2025appgen}
\bibfield{author}{\bibinfo{person}{Zihan Huang}, \bibinfo{person}{Tong Li}, {and} \bibinfo{person}{Yong Li}.} \bibinfo{year}{2025}\natexlab{}.
\newblock \showarticletitle{AppGen: Mobility-aware App Usage Behavior Generation for Mobile Users}.
\newblock \bibinfo{journal}{\emph{IEEE Transactions on Mobile Computing}} (\bibinfo{year}{2025}).
\newblock


\bibitem[Jimenez~Gutierrez et~al\mbox{.}(2024)]%
        {jimenez2024hipporag}
\bibfield{author}{\bibinfo{person}{Bernal Jimenez~Gutierrez}, \bibinfo{person}{Yiheng Shu}, \bibinfo{person}{Yu Gu}, \bibinfo{person}{Michihiro Yasunaga}, {and} \bibinfo{person}{Yu Su}.} \bibinfo{year}{2024}\natexlab{}.
\newblock \showarticletitle{Hipporag: Neurobiologically inspired long-term memory for large language models}.
\newblock \bibinfo{journal}{\emph{Advances in Neural Information Processing Systems}}  \bibinfo{volume}{37} (\bibinfo{year}{2024}), \bibinfo{pages}{59532--59569}.
\newblock


\bibitem[Ju et~al\mbox{.}(2025)]%
        {ju2025trajllm}
\bibfield{author}{\bibinfo{person}{Chenlu Ju}, \bibinfo{person}{Jiaxin Liu}, \bibinfo{person}{Shobhit Sinha}, \bibinfo{person}{Hao Xue}, {and} \bibinfo{person}{Flora Salim}.} \bibinfo{year}{2025}\natexlab{}.
\newblock \showarticletitle{Trajllm: A modular llm-enhanced agent-based framework for realistic human trajectory simulation}. In \bibinfo{booktitle}{\emph{Companion Proceedings of the ACM on Web Conference 2025}}. \bibinfo{pages}{2847--2850}.
\newblock


\bibitem[Latimer et~al\mbox{.}(2025)]%
        {latimer2025hindsight}
\bibfield{author}{\bibinfo{person}{Chris Latimer}, \bibinfo{person}{Nicol{\'o} Boschi}, \bibinfo{person}{Andrew Neeser}, \bibinfo{person}{Chris Bartholomew}, \bibinfo{person}{Gaurav Srivastava}, \bibinfo{person}{Xuan Wang}, {and} \bibinfo{person}{Naren Ramakrishnan}.} \bibinfo{year}{2025}\natexlab{}.
\newblock \showarticletitle{Hindsight is 20/20: Building Agent Memory that Retains, Recalls, and Reflects}.
\newblock \bibinfo{journal}{\emph{arXiv:2512.12818}} (\bibinfo{year}{2025}).
\newblock


\bibitem[Li et~al\mbox{.}(2025)]%
        {li2025memos}
\bibfield{author}{\bibinfo{person}{Zhiyu Li}, \bibinfo{person}{Chenyang Xi}, \bibinfo{person}{Chunyu Li}, \bibinfo{person}{Ding Chen}, \bibinfo{person}{Boyu Chen}, \bibinfo{person}{Shichao Song}, \bibinfo{person}{Simin Niu}, \bibinfo{person}{Hanyu Wang}, \bibinfo{person}{Jiawei Yang}, \bibinfo{person}{Chen Tang}, {et~al\mbox{.}}} \bibinfo{year}{2025}\natexlab{}.
\newblock \showarticletitle{Memos: A memory os for ai system}.
\newblock \bibinfo{journal}{\emph{arXiv:2507.03724}} (\bibinfo{year}{2025}).
\newblock


\bibitem[Maharana et~al\mbox{.}(2024)]%
        {maharana2024locomo}
\bibfield{author}{\bibinfo{person}{Adyasha Maharana}, \bibinfo{person}{Dong-Ho Lee}, \bibinfo{person}{Sergey Tulyakov}, \bibinfo{person}{Mohit Bansal}, \bibinfo{person}{Francesco Barbieri}, {and} \bibinfo{person}{Yuwei Fang}.} \bibinfo{year}{2024}\natexlab{}.
\newblock \showarticletitle{Evaluating very long-term conversational memory of llm agents}. In \bibinfo{booktitle}{\emph{The 62nd Annual Meeting of the Association for Computational Linguistics}}. \bibinfo{pages}{13851–13870}.
\newblock


\bibitem[MemU({[n.\,d.]})]%
        {memu}
MemU \bibinfo{year}{[n.\,d.]}\natexlab{}.
\newblock
\newblock
\shownote{https://github.com/NevaMind-AI/memU}.


\bibitem[Musto et~al\mbox{.}(2018)]%
        {musto2018framework}
\bibfield{author}{\bibinfo{person}{Cataldo Musto}, \bibinfo{person}{Giovanni Semeraro}, \bibinfo{person}{Cosimo Lovascio}, \bibinfo{person}{Marco De~Gemmis}, {and} \bibinfo{person}{Pasquale Lops}.} \bibinfo{year}{2018}\natexlab{}.
\newblock \showarticletitle{A framework for holistic user modeling merging heterogeneous digital footprints}. In \bibinfo{booktitle}{\emph{Adjunct Publication of the 26th Conference on User Modeling, Adaptation and Personalization}}. \bibinfo{pages}{97--101}.
\newblock


\bibitem[Naiman et~al\mbox{.}(2023)]%
        {naiman2023generative}
\bibfield{author}{\bibinfo{person}{Ilan Naiman}, \bibinfo{person}{N~Benjamin Erichson}, \bibinfo{person}{Pu Ren}, \bibinfo{person}{Michael~W Mahoney}, {and} \bibinfo{person}{Omri Azencot}.} \bibinfo{year}{2023}\natexlab{}.
\newblock \showarticletitle{Generative modeling of regular and irregular time series data via koopman VAEs}.
\newblock \bibinfo{journal}{\emph{arXiv:2310.02619}} (\bibinfo{year}{2023}).
\newblock


\bibitem[{OpenAI}(2025a)]%
        {openai_gpt51mini}
\bibfield{author}{\bibinfo{person}{{OpenAI}}.} \bibinfo{year}{2025}\natexlab{a}.
\newblock \bibinfo{title}{{GPT-5.1 mini}}.
\newblock \bibinfo{howpublished}{\url{https://openai.com}}.
\newblock


\bibitem[{OpenAI}(2025b)]%
        {openai_textembedding}
\bibfield{author}{\bibinfo{person}{{OpenAI}}.} \bibinfo{year}{2025}\natexlab{b}.
\newblock \bibinfo{title}{{text-embedding-3-small}}.
\newblock \bibinfo{howpublished}{\url{https://openai.com}}.
\newblock


\bibitem[Park et~al\mbox{.}(2023)]%
        {park2023generative_agents}
\bibfield{author}{\bibinfo{person}{Joon~Sung Park}, \bibinfo{person}{Joseph~C. O'Brien}, \bibinfo{person}{Carrie~J. Cai}, \bibinfo{person}{Meredith~Ringel Morris}, \bibinfo{person}{Percy Liang}, {and} \bibinfo{person}{Michael~S. Bernstein}.} \bibinfo{year}{2023}\natexlab{}.
\newblock \showarticletitle{Generative Agents: Interactive Simulacra of Human Behavior}. In \bibinfo{booktitle}{\emph{Proceedings of the 36th Annual ACM Symposium on User Interface Software and Technology}}. \bibinfo{pages}{1--22}.
\newblock


\bibitem[Rasmussen et~al\mbox{.}(2025)]%
        {rasmussen2025zep}
\bibfield{author}{\bibinfo{person}{Preston Rasmussen}, \bibinfo{person}{Pavlo Paliychuk}, \bibinfo{person}{Travis Beauvais}, \bibinfo{person}{Jack Ryan}, {and} \bibinfo{person}{Daniel Chalef}.} \bibinfo{year}{2025}\natexlab{}.
\newblock \showarticletitle{Zep: a temporal knowledge graph architecture for agent memory}.
\newblock \bibinfo{journal}{\emph{arXiv:2501.13956}} (\bibinfo{year}{2025}).
\newblock


\bibitem[Salama et~al\mbox{.}(2025)]%
        {salama2025meminsight}
\bibfield{author}{\bibinfo{person}{Rana Salama}, \bibinfo{person}{Jason Cai}, \bibinfo{person}{Michelle Yuan}, \bibinfo{person}{Anna Currey}, \bibinfo{person}{Monica Sunkara}, \bibinfo{person}{Yi Zhang}, {and} \bibinfo{person}{Yassine Benajiba}.} \bibinfo{year}{2025}\natexlab{}.
\newblock \showarticletitle{{M}em{I}nsight: Autonomous Memory Augmentation for {LLM} Agents}. In \bibinfo{booktitle}{\emph{Proceedings of the 2025 Conference on Empirical Methods in Natural Language Processing}}. \bibinfo{pages}{33136--33152}.
\newblock


\bibitem[Squire(2004)]%
        {squire2004memory}
\bibfield{author}{\bibinfo{person}{Larry~R Squire}.} \bibinfo{year}{2004}\natexlab{}.
\newblock \showarticletitle{Memory systems of the brain: a brief history and current perspective}.
\newblock \bibinfo{journal}{\emph{Neurobiology of learning and memory}} \bibinfo{volume}{82}, \bibinfo{number}{3} (\bibinfo{year}{2004}), \bibinfo{pages}{171--177}.
\newblock


\bibitem[Tan et~al\mbox{.}(2023)]%
        {tan2023timelineqa}
\bibfield{author}{\bibinfo{person}{Wang-Chiew Tan}, \bibinfo{person}{Jane Dwivedi-Yu}, \bibinfo{person}{Yuliang Li}, \bibinfo{person}{Lambert Mathias}, \bibinfo{person}{Marzieh Saeidi}, \bibinfo{person}{Jing~Nathan Yan}, {and} \bibinfo{person}{Alon~Y. Halevy}.} \bibinfo{year}{2023}\natexlab{}.
\newblock \showarticletitle{{T}imeline{QA}: A Benchmark for Question Answering over Timelines}. In \bibinfo{booktitle}{\emph{Findings of the Association for Computational Linguistics: ACL 2023}}. \bibinfo{pages}{77--91}.
\newblock


\bibitem[The Lifelog Search Challenge(2026)]%
        {lifelog}
The Lifelog Search Challenge \bibinfo{year}{2018-2026}\natexlab{}.
\newblock
\newblock
\shownote{https://lifelogsearch.org/lsc/}.


\bibitem[Tran et~al\mbox{.}(2025)]%
        {tran2025openlifelogqa}
\bibfield{author}{\bibinfo{person}{Quang-Linh Tran}, \bibinfo{person}{Binh Nguyen}, \bibinfo{person}{Gareth~JF Jones}, {and} \bibinfo{person}{Cathal Gurrin}.} \bibinfo{year}{2025}\natexlab{}.
\newblock \showarticletitle{OpenLifelogQA: An Open-Ended Multi-Modal Lifelog Question-Answering Dataset}.
\newblock \bibinfo{journal}{\emph{arXiv:2508.03583}} (\bibinfo{year}{2025}).
\newblock


\bibitem[Wan et~al\mbox{.}(2017)]%
        {wan2017variational}
\bibfield{author}{\bibinfo{person}{Zhiqiang Wan}, \bibinfo{person}{Yazhou Zhang}, {and} \bibinfo{person}{Haibo He}.} \bibinfo{year}{2017}\natexlab{}.
\newblock \showarticletitle{Variational autoencoder based synthetic data generation for imbalanced learning}. In \bibinfo{booktitle}{\emph{2017 IEEE symposium series on computational intelligence (SSCI)}}. \bibinfo{pages}{1--7}.
\newblock


\bibitem[Wang and Chen(2025)]%
        {wang2025mirix}
\bibfield{author}{\bibinfo{person}{Yu Wang} {and} \bibinfo{person}{Xi Chen}.} \bibinfo{year}{2025}\natexlab{}.
\newblock \showarticletitle{Mirix: Multi-agent memory system for llm-based agents}.
\newblock \bibinfo{journal}{\emph{arXiv:2507.07957}} (\bibinfo{year}{2025}).
\newblock


\bibitem[Wu et~al\mbox{.}(2025)]%
        {wu2025longmemeval}
\bibfield{author}{\bibinfo{person}{Di Wu}, \bibinfo{person}{Hongwei Wang}, \bibinfo{person}{Wenhao Yu}, \bibinfo{person}{Yuwei Zhang}, \bibinfo{person}{Kai-Wei Chang}, {and} \bibinfo{person}{Dong Yu}.} \bibinfo{year}{2025}\natexlab{}.
\newblock \showarticletitle{LongMemEval: Benchmarking Chat Assistants on Long-Term Interactive Memory}. In \bibinfo{booktitle}{\emph{The Thirteenth International Conference on Learning Representations}}.
\newblock


\bibitem[Xiong et~al\mbox{.}(2023)]%
        {xiong2023trajsgan}
\bibfield{author}{\bibinfo{person}{Gang Xiong}, \bibinfo{person}{Zhishuai Li}, \bibinfo{person}{Meihua Zhao}, \bibinfo{person}{Yu Zhang}, \bibinfo{person}{Qinghai Miao}, \bibinfo{person}{Yisheng Lv}, {and} \bibinfo{person}{Fei-Yue Wang}.} \bibinfo{year}{2023}\natexlab{}.
\newblock \showarticletitle{TrajSGAN: A semantic-guiding adversarial network for urban trajectory generation}.
\newblock \bibinfo{journal}{\emph{IEEE Transactions on Computational Social Systems}} \bibinfo{volume}{11}, \bibinfo{number}{2} (\bibinfo{year}{2023}), \bibinfo{pages}{1733--1743}.
\newblock


\bibitem[Xu et~al\mbox{.}({[n.\,d.]})]%
        {xu2022beyond}
\bibfield{author}{\bibinfo{person}{Jing Xu}, \bibinfo{person}{Arthur Szlam}, {and} \bibinfo{person}{Jason Weston}.} \bibinfo{year}{[n.\,d.]}\natexlab{}.
\newblock \showarticletitle{Beyond Goldfish Memory: Long-Term Open-Domain Conversation}. In \bibinfo{booktitle}{\emph{Proceedings of the 60th Annual Meeting of the Association for Computational Linguistics}}, \bibfield{editor}{\bibinfo{person}{Smaranda Muresan}, \bibinfo{person}{Preslav Nakov}, {and} \bibinfo{person}{Aline Villavicencio}} (Eds.). \bibinfo{pages}{5180--5197}.
\newblock


\bibitem[Xu et~al\mbox{.}(2025)]%
        {xu2025a-mem}
\bibfield{author}{\bibinfo{person}{Wujiang Xu}, \bibinfo{person}{Zujie Liang}, \bibinfo{person}{Kai Mei}, \bibinfo{person}{Hang Gao}, \bibinfo{person}{Juntao Tan}, {and} \bibinfo{person}{Yongfeng Zhang}.} \bibinfo{year}{2025}\natexlab{}.
\newblock \showarticletitle{A-mem: Agentic memory for llm agents}.
\newblock \bibinfo{journal}{\emph{Advances in Neural Information Processing Systems}} (\bibinfo{year}{2025}).
\newblock


\bibitem[Zacks and Tversky(2001)]%
        {zacks2001event}
\bibfield{author}{\bibinfo{person}{Jeffrey~M Zacks} {and} \bibinfo{person}{Barbara Tversky}.} \bibinfo{year}{2001}\natexlab{}.
\newblock \showarticletitle{Event structure in perception and conception.}
\newblock \bibinfo{journal}{\emph{Psychological bulletin}} \bibinfo{volume}{127}, \bibinfo{number}{1} (\bibinfo{year}{2001}), \bibinfo{pages}{3}.
\newblock


\bibitem[Zhang(2024)]%
        {zhang2024guided}
\bibfield{author}{\bibinfo{person}{Jiarui Zhang}.} \bibinfo{year}{2024}\natexlab{}.
\newblock \showarticletitle{Guided Profile Generation Improves Personalization with Large Language Models}. In \bibinfo{booktitle}{\emph{Findings of the Association for Computational Linguistics: EMNLP}}. \bibinfo{pages}{4005--4016}.
\newblock


\bibitem[Zhang et~al\mbox{.}(2017)]%
        {zhang2017privbayes}
\bibfield{author}{\bibinfo{person}{Jun Zhang}, \bibinfo{person}{Graham Cormode}, \bibinfo{person}{Cecilia~M Procopiuc}, \bibinfo{person}{Divesh Srivastava}, {and} \bibinfo{person}{Xiaokui Xiao}.} \bibinfo{year}{2017}\natexlab{}.
\newblock \showarticletitle{Privbayes: Private data release via bayesian networks}.
\newblock \bibinfo{journal}{\emph{ACM Transactions on Database Systems (TODS)}} \bibinfo{volume}{42}, \bibinfo{number}{4} (\bibinfo{year}{2017}), \bibinfo{pages}{1--41}.
\newblock


\bibitem[Zhong et~al\mbox{.}(2024)]%
        {zhong2024memorybank}
\bibfield{author}{\bibinfo{person}{Wanjun Zhong}, \bibinfo{person}{Lianghong Guo}, \bibinfo{person}{Qiqi Gao}, \bibinfo{person}{He Ye}, {and} \bibinfo{person}{Yanlin Wang}.} \bibinfo{year}{2024}\natexlab{}.
\newblock \showarticletitle{Memorybank: Enhancing large language models with long-term memory}. In \bibinfo{booktitle}{\emph{Proceedings of the AAAI Conference on Artificial Intelligence}}, Vol.~\bibinfo{volume}{38}. \bibinfo{pages}{19724--19731}.
\newblock


\end{thebibliography}

\appendix

\section{Related Work on Behavior Modeling}

Beyond memory benchmarking, our research is closely situated within the broader field of human behavior modeling. The idea of organizing personal information traces back to the Memex system \cite{bush1945we}, and has since been advanced through projects such as MyLifeBits \cite{gemmell2006mylifebits}, Musto \cite{musto2018framework}, and Cyber-I \cite{guo2018user}, among other notable lifelogging research \cite{lifelog,tran2025openlifelogqa}. MyLifeBits \cite{gemmell2006mylifebits} utilized simple SQL-based architectures and keyword indexing to manage personal information. Musto \cite{musto2018framework} proposed an integrated user modeling framework that fuses heterogeneous digital footprints from social networks and personal devices, while Cyber-I \cite{guo2018user} extended this idea by modeling affective, behavioral, and cognitive dimensions. The Lifelog Search Challenge (LSC) \cite{lifelog,tran2025openlifelogqa} further explored records of personal events generated from  digital sources such as photos or shopping data. Despite these advances, real-world lifelog data remain limited in scale and availability due to privacy concerns, ethical limitations, and prohibitive costs.

Synthetic data generation has emerged as a promising approach to address these challenges for modeling human behaviors. Early work relied on statistical methods~\cite{zhang2017privbayes} and generative models such as GANs~\cite{xiong2023trajsgan} and VAEs~\cite{wan2017variational} ~\cite{naiman2023generative}. More recently, large language models (LLMs) have enabled the creation of rich textual and structured personal data~\cite{zhang2024guided, ju2025trajllm}. However, these studies mainly focus on synthesizing single-aspect data, such as user profiles~\cite{zhang2024guided}, urban mobility traces~\cite{xiong2023trajsgan,ju2025trajllm}, or simple app usage trajectories \cite{huang2025appgen}, typically lacking user-centric full memory modeling. Recent works, including TimelineQA ~\cite{tan2023timelineqa} and Generative Agents ~\cite{park2023generative_agents}, partially address this issue but struggle with data fidelity. These approaches either do not ground on realistic temporal and location information ~\cite{park2023generative_agents} or rely on rigid template-based generation ~\cite{tan2023timelineqa}, failing to reflect the complexity of actual human experience.

\section{Scalability}
\label{app:scale}
Parallel thematic event decomposition is provided in Algorithm 1.
\begin{algorithm}[]
    \caption{Parallel Event Decomposition (1-Day Threshold)}
    \label{alg:core_event_decomposition}
        \begin{algorithmic}[1]
    \REQUIRE $EventTree$ with all thematic events as leaves, Time threshold $\tau=1$ day
\ENSURE Expanded $EventTree$ with partonomic hierarchy 

\STATE \textbf{Function} $\text{DFSParallelEventDecomp}(EventTree, \tau)$
    \STATE $dfs\_stack \leftarrow \text{Empty Stack}$
    \STATE Assign all leaf nodes with $splittable = True$
    
    \FOR{each leaf $node$ in $EventTree$ where $node.splittable = True$}
        \STATE $\text{Push}(dfs\_stack, node)$
    \ENDFOR
    
    \WHILE{$dfs\_stack$ is not Empty}
        \STATE $current\_node \leftarrow \text{Pop}(dfs\_stack)$
        
        \STATE Assign $current\_node$ to an available worker
        \STATE \textit{\textbf{The following is done in parallel for all popped nodes in current batch}}
        \STATE $subevts \leftarrow \text{LLM}_{\text{decomp}}(current\_node, \tau)$;
        
        \FOR{each $sub_e \in subevts$}
            \IF {$sub_e.duration < \tau$}
                \STATE $sub_e.splittable = False$
            \ELSE
                \STATE $sub_e.splittable = True$
                \STATE $\text{Push}(dfs\_stack, sub_e)$
            \ENDIF
        \ENDFOR
        
        \STATE $current\_node.subevents = subevts$
    \ENDWHILE
    
    \STATE RETURN $EventTree$
\end{algorithmic}
\end{algorithm}

\section{Dataset Statistics}
The statistical data of the dataset can be found in Table \ref{tab:dataset_stats}.

\begin{table}[t]
\centering
\caption{Dataset statistics}
\label{tab:dataset_stats}
\begin{tabular}{@{}lr@{}}
\toprule
\textbf{Metric} & \textbf{Value} \\
\midrule
Total users & 10 \\
Duration per user & 1 year \\
Total events & 51491 \\
Events per user (mean) & 5149 \\
Events per user (median) & 5147 \\
\midrule
Persona tokens (mean) & 17730.8\\
Realtionship (mean) & 24.4\\
location (mean) & 15\\
\midrule
Contacts per user (mean) & 21.9 \\
Calls per user (mean) & 1040.4 \\
SMS messages per user (mean) & 1812.9 \\
Calendar events per user (mean) & 234.4 \\
Agent chat conversations per user & 688.5 \\
Photos per user (mean) & 1233.4 \\
Notes per user (mean) & 363.1 \\
Push notifications per user (mean) & 2350.3 \\
Fitness health records per user & 301.0 \\
\midrule
Monthly summaries per user & 12 \\
Summary length (words, per user) & 1737.7\\
\midrule
Context depth (tokens) per user & 3.66M \\
Total dataset size & 332 MB \\
Generation time per user & 8 hours \\
\bottomrule
\end{tabular}
\end{table}

\section{Detailed Evaluation Metrics}
\label{app:evaluation}
\subsection{Manual Evaluation}
\label{app:manual evaluation}

\paragraph{Scoring Criteria for Rationality Evaluation:}
\begin{enumerate}
 \item [(1)]For each piece of daily activity data, we assessed the rationality of its temporal coherence, content, location, and participant. If any of these dimensions was deemed unreasonable, the data was marked as invalid, and the proportion of invalid data was calculated at last.

 \item [(2)]For each piece of phone data, we traced it back to the corresponding daily activity data in the matching time period, then evaluated the consistency between phone operations and daily activities as well as the rationality of phone operation content. If either dimension was unreasonable, the data was labeled as invalid, with the final proportion of invalid data statistically tabulated.
\end{enumerate}

\paragraph{Scoring Criteria for Diversity Evaluation:}
\begin{enumerate}
    \item [(1)] There are significant differences between the daily activities of the subject and those of other subjects, which can reflect the individual's unique characteristics (2 points).
    \item [(2)] There are significant differences in the activities of the same subject on different dates, and the activities are not homogeneous (2 points).
    \item [(3)] On the basis of reflecting individual characteristics, the data includes a certain number of randomly diverse events and population common events, rather than only events that follow the stereotypes of character portraits (2 points).
    \item [(4)] On the basis of criteria (1) and (2), the fine-grained metrics of the subject (e.g., sleep duration, body weight, and dietary intake) also exhibit obvious fluctuations (2 points).
    \item [(5)] Verified by manual analysis, the data can reflect the complexity, unpredictability and rich diversity of real life, and the overall content is coherent and without sense of incongruity (2 points).
\end{enumerate}

\noindent \textbf{Total Score}: 10 points (2 points are assigned to each of the 5 evaluation criteria,0 points for completely unsatisfactory, 1 point for relatively satisfactory, and 2 points for highly satisfactory.).

\begin{figure}[t]
\centering
\includegraphics[width=0.9\linewidth]{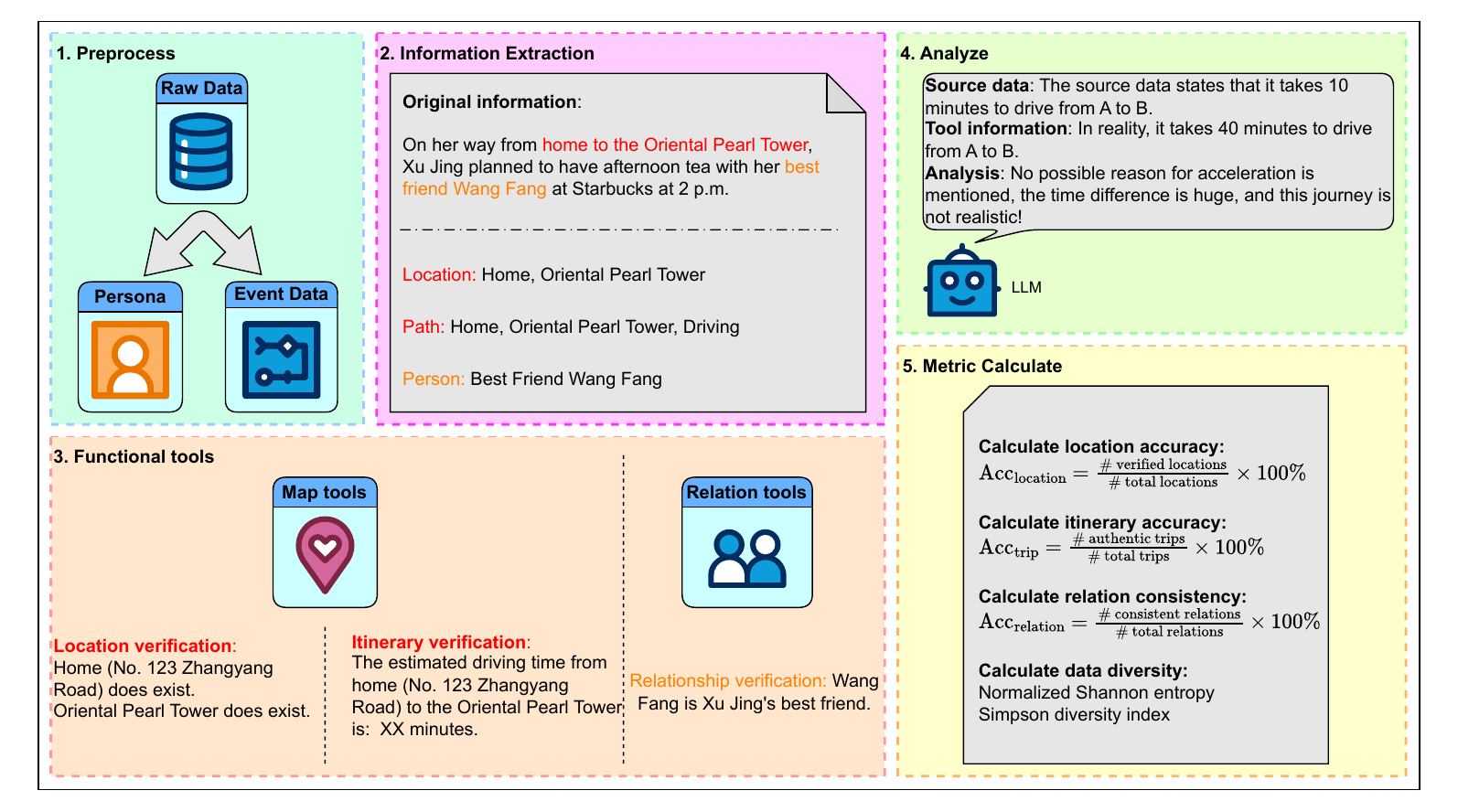}
\caption{Automated data quality evaluation pipeline for consistency, authenticity, and diversity assessment.}
\label{fig:quality_eval}
\end{figure}

\subsection{Automatic Evaluation Metrics}
We implement a comprehensive and fully automated data quality evaluation pipeline (Figure~\ref{fig:quality_eval}) to assess the \textbf{consistency}, \textbf{authenticity}, and \textbf{diversity} of the generated event data. Overall, the proposed framework consists of two complementary components:

\begin{itemize}
    \item \textit{Intra-source consistency evaluation}, which verifies whether generated events remain coherent with the original structured source data, such as persona profiles and predefined social relations. This component primarily relies on lookup-table construction and LLM-assisted information extraction to determine whether referenced entities and relations can be consistently traced back to the source data.
    
    \item \textit{Extra-source authenticity evaluation}, which validates generated content against real-world knowledge beyond the source data. In particular, we leverage real-world geographic and transportation information provided by the map API to assess the plausibility of locations and travel trajectories mentioned in the generated events.
\end{itemize}

Based on the above two components, we evaluate the quality of generated events from four complementary aspects: \textit{relation consistency}, \textit{location authenticity}, \textit{trip authenticity}, and \textit{event diversity}.

\paragraph{Relation Consistency}

To evaluate character and relationship consistency, we construct a relation lookup dictionary from persona profiles, which specifies the expected social network for each character. We then apply LLM-based information extraction to identify character mentions and relational references in generated event narratives.

If a character appears in an event but cannot be matched to any entry in the predefined relation dictionary, the mention is marked as inconsistent. Relation consistency accuracy is computed as:
\[
\text{Acc}_{\text{person}} =
\frac{\#\ \text{consistent relation mentions}}
{\#\ \text{total relation mentions}} \times 100\%
\]

This metric quantifies the degree to which generated events adhere to the original interpersonal structure.

\paragraph{Location Authenticity}

For location authenticity, place names are automatically extracted from event descriptions and queried against the AMap API. Locations that can be successfully verified by the map service are labeled as authentic, while unverifiable or ambiguous locations are considered invalid.

Location authenticity accuracy is calculated as:
\[
\text{Acc}_{\text{location}} =
\frac{\#\ \text{verified locations}}
{\#\ \text{total extracted locations}} \times 100\%
\]

This evaluation ensures that generated locations correspond to real-world geographic entities.

\paragraph{Trip Authenticity}

For events involving travel or movement, we further evaluate trip-level authenticity. Given extracted origin--destination pairs and temporal information from event narratives, we query the AMap API to obtain reference travel durations and routes.

A synthesized trip is considered authentic if its temporal deviation falls within a predefined tolerance threshold (±20\% relative deviation or ±30 minutes absolute deviation) compared to the API-provided reference. Trips exceeding this threshold are still considered authentic if the original data provides a plausible explanation for the temporal discrepancy (e.g., traffic congestion). Trip authenticity accuracy is defined as:
\[
\text{Acc}_{\text{trip}} =
\frac{\#\ \text{authentic trips}}
{\#\ \text{total evaluated trips}} \times 100\%
\]

This metric assesses whether generated travel descriptions are consistent with realistic transportation constraints.

\paragraph{Diversity Evaluation}

To quantify event diversity, we analyze the distribution of event types using two complementary metrics.

Normalized Shannon Entropy measures the uncertainty and coverage of event types:
\[
H = -\sum_{i=1}^{K} p_i \log p_i, \quad
H_{\text{norm}} = \frac{H}{\log K}
\]
where $K$ denotes the total number of event categories and $p_i$ is the proportion of events belonging to category $i$.

Simpson Diversity Index captures the evenness of the distribution. We first compute the concentration measure:
\[
D = \sum_{i=1}^{K} p_i^2
\]
and define the Simpson diversity index as:
\[
\text{Simpson} = 1 - D = 1 - \sum_{i=1}^{K} p_i^2
\]

Higher values of both metrics indicate more balanced and diverse event distributions, whereas lower values suggest concentration in a small number of event types.

\section{Additional Results}
\begin{figure}[t]
\centering
\includegraphics[width=0.4\textwidth, height=3cm]{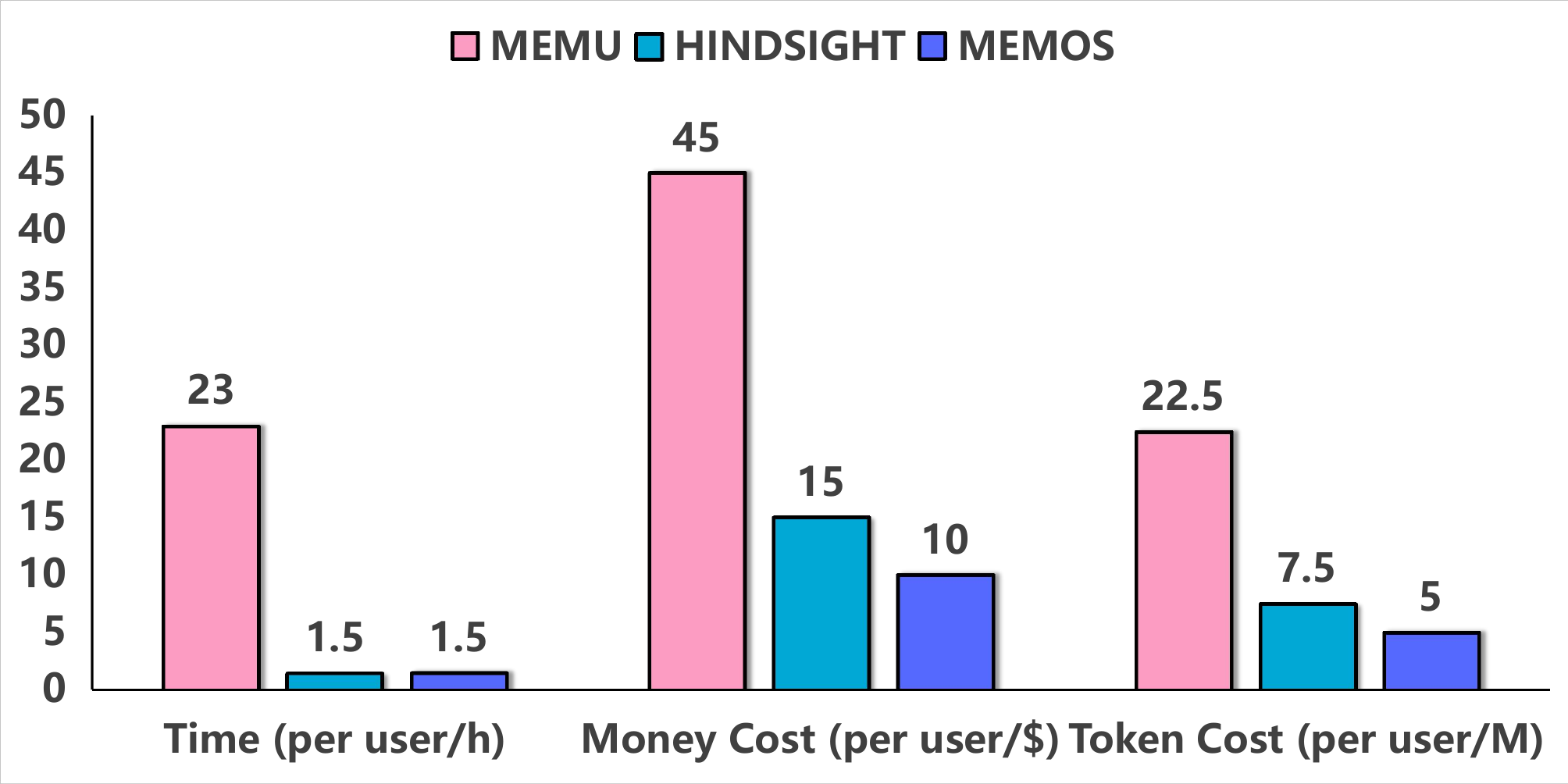}
\caption{Time and cost of each system.}
\label{fig:time_and_cost}
\end{figure}

\subsection{Time Complexity and Cost Analysis}
We conducted a coarse-grained analysis and estimation of the time and cost required by each system to process a single user, as summarized in Figure~\ref{fig:time_and_cost}. It is evident that MEMU incurs substantially higher time overhead and monetary cost than Hindsight and MemOS. This difference primarily stems from MEMU’s reliance on frequent LLM invocations throughout its pipeline. Beyond text embedding, MEMU invokes the LLM at nearly every processing stage, with a single memory operation requiring multiple LLM calls. In contrast, Hindsight and MemOS support batch processing and avoid repeated LLM invocations during memory handling, resulting in significantly lower computational overhead and cost.



















\subsection{Extension of Error Analysis}
\label{errorcase}
\subsubsection{Failure to retrieve correct evidence}
\begin{itemize}
    \item \textit{Question:} After Ye Mingxuan started his weight loss plan in May, did he perform his first aerobic exercise from the plan before or after his first strength training session?

    \item \textit{Generated Answer from MemU:} Uncertain: There is a strength training record from May  (around May 17), but there is no clear first aerobic record from May for comparison.

    \item \textit{Standard Answer:} After.

    \item \textit{Analysis:} The system's memory clearly recorded that Ye Mingxuan first performed weight training and then aerobic training, but the system did not realize that weight training was strength training, which is why it could not retrieve the specific content.
\end{itemize}
\subsubsection{Incomplete evidence retrieval}
\begin{itemize}
    \item \textit{Question:} What specific tasks did Feng Haoran complete in February in preparation for his promotion performance review report?

    \item \textit{Generated Answer from MemU:} February work: Organize annual learning materials; refine the "Teamwork and Leadership" section; draft a six-month learning plan; send the initial draft of the performance review to General Manager Qian and set a schedule for optimization.

    \item \textit{Standard Answer:} He conceived the report framework, organized team collaboration materials, compiled personal continuous learning documentation, wrote the section on team contributions, refined the future learning plan, and integrated the findings to complete the first draft of the report.

    \item \textit{Analysiz:} The answer is incomplete. The system's memory records that a report framework was conceived, team collaboration materials were organized, and personal continuous learning documents were compiled, but these were not included in the answer.
\end{itemize}
\subsubsection{Hallucinated content}
\begin{itemize}
    \item \textit{Question:} What are the specific dates and locations for Yu Xiaowen's high school reunion (initiated by Chen Siyu) on May 30, 2025?

    \item \textit{Generated Answer from MemU:} May 30th entry: Scheduled to return to Luoyang on June 1st, 2025 for a classmate reunion; another reunion on June 18th, 2025 at Dian Dou De Restaurant (Chen Siyu will be present).

    \item \textit{Standard Answer:} Unable to answer.

    \item \textit{Analysis:} The data only showed that a gathering took place but did not specify when the next gathering would take place, leading to over-inference in the system's response.
\end{itemize}
\subsubsection{Incorrect reasoning}
\begin{itemize}
    \item \textit{Question:} Where did Feng Haoran mainly work in January?

    \item \textit{Generated Answer from MemU:} Primarily based in the company office in Changsha; during holidays or business trips, work remotely or via video from hotels or homes.

    \item \textit{Standard Answer:} Huatai Securities Co., Ltd. has its office located at No. 88, Furong Middle Road, Kaifu District, Changsha.

    \item \textit{Analysis:} The system ignored the crucial limiting information that Feng Haoran worked at the company throughout January, and only in later months did he work in more locations.
\end{itemize}
\subsubsection{Memory omission}
\begin{itemize}
    \item \textit{Question:} During the Spring Festival, while helping his mother prepare ingredients, Feng Haoran had an in-depth discussion with his father about which report?

    \item \textit{Generated Answer from MemU:} The provided materials did not specify which report it was, and the information was insufficient to determine the source.

    \item \textit{Standard Answer:} Central Bank Monetary Policy Report.

    \item \textit{Analysis:} The system considered the Central Bank Monetary Policy Report information unimportant and did not fully incorporate it into its memory, merely using "report" as a substitute. This resulted in a lack of memorized information when answering the question, leading to an inability to respond correctly.
\end{itemize}

\subsubsection{Hallucination in Memory Unit Construction}
\begin{itemize}
    \item \textit{Question:} In July, when Ma Xiulan asked her neighbor for advice on aphid control, what two homemade remedies did the neighbor recommend?

    \item \textit{Generated Answer from MemOS:} Tobacco water and chili-garlic water
    \item \textit{Standard Answer:} Tobacco water and laundry detergent water.
    \item \textit{Analysis:} We analyzed the corresponding memory units of Memos and found that it did capture the note data recording tobacco water and laundry detergent water. However, it autonomously added the non-existent hallucination of chili-garlic water—which was not present in the original data—during the process of generating summaries for memory units, leading to an error.
\end{itemize}

\subsubsection{Retrieval without Temporal Constraints}
\begin{itemize}
    \item \textit{Question:} When handling the annual fixed expenses due in August, what specific arrangements did I make on the day before the actual payment?

    \item \textit{Generated Answer from Hindsight:} There is no explicit record in the memory of the specific arrangements you made on "the day before the actual payment" (i.e., August 10, 2025).
    \item \textit{Standard Answer:} I withdrew 600 yuan in cash from the family emergency cash box, retrieved the house property certificate and household registration booklet from the chest of drawers, put them into a kraft paper document bag, and placed the bag in a prominent position on the TV cabinet in the living room.
    \item \textit{Analysis:} We found that the payment event was successfully retrieved; however, the retrieval purely based on similarity introduced irrelevant financial payment-related information for the entire year that was unrelated to the query target, and failed to retrieve events from the day before the payment based on time constraints, resulting in the inability to answer the query.
\end{itemize}


\section{Impact and Use Cases}
\label{sec:discussion}

\subsection{Implications for Memory System Design}

Our benchmark results reveal several insights for designing effective memory-augmented personal agents:

\textbf{Comprehensive Memory Structure Modeling}
Developing a comprehensive system that addresses the multiple dimensions of human memory—such as declarative and non-declarative memory—is vital for realizing truly personalized AI agents. While frameworks like Hindsight and MemOS have addressed to some extent, a system that holistically addresses all aspects of \ourbenchmark{} has yet to be developed.


\textbf{Temporal Reasoning and Knowledge Updating Are a Major Bottleneck.}  Current systems rely primarily on similarity-based retrieval and lack the temporal reasoning necessary to handle time-sensitive data. This limitation introduces significant noise and can lead to the omission of critical information. To address this, it is recommended to implement specialized temporal indexing and time-aware query processing.


\textbf{Multi-source integration remains challenging.} Systems that performed exceptionally well on previous benchmarks consisting solely of chat-based data exhibit a significant performance decline on ours. This indicates that integrating and analyzing multi-type data to infer personal life patterns poses considerable challenges, and future systems need to optimize the organization and reasoning of multi-source information. 



\subsection{Enabling Research Directions}

The \ourbenchmark{} dataset and benchmark enable systematic investigation of numerous research questions:

\textbf{Memory consolidation and forgetting.} How should systems decide what to remember, summarize, or forget? Our dataset presents challenges of a long time span and a large volume of long texts.

\textbf{Personalized memory organization.} Do optimal memory structures vary by persona (e.g., detail-oriented vs. big-picture individuals)? Our diverse personas enable persona-conditioned memory research.

\textbf{Proactive memory retrieval.} Beyond reactive QA, how can systems proactively surface relevant memories? Our rich event context supports evaluation of memory-triggered interventions.

\textbf{Privacy-preserving memory.} What information can be safely retained vs. summarized for privacy? Our synthetic data allows experimentation without real privacy risks.

\subsection{Application Scenarios}

\textbf{Personal AI assistants.} Agents that remember user preferences, track commitments, and provide personalized recommendations based on long-term behavioral patterns.

\textbf{Digital health coaching.} Systems that monitor health trends, correlate lifestyle factors with outcomes, and provide personalized interventions grounded in individual history.

\textbf{Behavioral research.} Controlled experimentation on human behavior modeling, habit formation, and lifestyle dynamics without privacy-sensitive real data.

\textbf{Memory-augmented LLM development.} Standardized benchmark for comparing memory architectures, retrieval strategies, and summarization approaches.

\section{Limitations and Ethical Considerations}
\label{sec:ethics}

\subsection{Limitations}

\textbf{Synthetic Realism Boundaries.} Despite our efforts to ensure coherence and diversity, synthetic data cannot fully capture the complexity, unpredictability, and rich contextual nuances of real human lives. Edge cases, rare events, and genuine emotional depth remain challenging to synthesize convincingly.

\textbf{Coverage Gaps.} Our dataset focuses on individuals in Chinese adult population with mobile phones. It does not represent all demographics, cultures, or lifestyles. Users with disabilities, non-standard living situations, or marginalized communities may be underrepresented.

\textbf{Persona and Event Assumptions.} Persona generation relies on anonymized survey data and social priors that may encode societal biases. Event synthesis uses LLM-based generation, which inherits biases from training data. We attempt to mitigate this through diversity controls and manual review, but systematic biases may remain.

\textbf{Health Data Disclaimers.} Health and sports records are \textit{illustrative and non-clinical}. They should not be used for medical research, diagnosis, or treatment recommendations. The data represent simplified tracking signals, not comprehensive medical records.

\textbf{Evaluation Limitations.} Open-ended QA scoring relies on LLM judges, which may introduce subjectivity and variance. Scoring points are designed to be objective, but edge case interpretation may vary. Multiple-choice questions are more objective but limited in capturing nuanced reasoning.

\subsection{Ethical Considerations}

\textbf{Privacy and Consent.} Although our dataset is entirely synthetic, it represents realistic personal scenarios that could inadvertently resemble real individuals. We take precautions: (1) all personas are composites from anonymized aggregates, not copies of real people; (2) sensitive attributes are removed or transformed; (3) generation is seeded randomly to prevent reidentification.

\textbf{Misuse Risks.} Realistic personal data, even synthetic, could be misused for:
\begin{itemize}
    \item Training surveillance or manipulation systems
    \item Generating deceptive personas for social engineering
    \item Creating plausible fake identities for fraud
\end{itemize}

We mitigate these risks by: (1) releasing data under a license prohibiting harmful use; (2) providing clear documentation of synthetic origins; (3) encouraging responsible use for research and beneficial applications only. However, we cannot fully prevent misuse. We rely on community norms and encourage dataset users to consider ethical implications of their work.

\textbf{Health-Related Cautions.} Health data in our dataset are \textit{not medical-grade}. They represent simplified activity tracking for lifestyle context. Any system trained on this data should \textit{not} be deployed for clinical decision-making without extensive validation on real medical data and regulatory approval.

\textbf{Informed Use.} We commit to transparency:
\begin{itemize}
    \item Comprehensive documentation of generation process, assumptions, and limitations
    \item Clear labeling as synthetic data in all releases
    \item Ongoing engagement with the research community to address emerging ethical concerns
\end{itemize}

\subsection{Responsible Use Recommendations}
The code repository for this study is released under the Apache License 2.0.
We recommend that researchers and practitioners using \ourbenchmark{}:
\begin{enumerate}
    \item Clearly disclose use of synthetic data in publications and deployments
    \item Validate findings on real data before drawing strong conclusions
    \item Consider demographic and cultural limitations when generalizing results
    \item Implement additional safeguards (bias audits, privacy protections) when building systems on this data
    \item Adhere to ethical guidelines for AI research and deployment in their respective domains
\end{enumerate}

\section{Dataset Schema Details}
\label{app:schema}

\subsection{Persona Schema}

\begin{lstlisting}
{
  "name": "Yu Xiaowei",
  "birth": "1986-03-14",
  "age": 35,
  "nationality": "Han",
  "home_address": {
    "province": "Hong Kong Special Administrative Region",
    "city": "Hong Kong",
    "district": "Kowloon City District",
    "street_name": "Prince Edward Road West",
    "street_number": "No. 328"
  },
  "birth_place": {
    "province": "Hong Kong Special Administrative Region",
    "city": "Hong Kong",
    "district": "Kowloon City District"
  },
  "gender": "Female",
  "education": "General Senior High School",
  "job": "Insurance Agent",
  "occupation": "AIA Hong Kong",
  "workplace": {
    "province": "Hong Kong Special Administrative Region",
    "city": "Hong Kong",
    "district": "Wan Chai District",
    "street_name": "Gloucester Road",
    "street_number": "AIA Central, No. 138"
  },
  "belief": "No religious belief",
  "salary": 300000.0,
  "body": {
    "height": 155,
    "weight": 55.0,
    "BMI": 22.89
  },
  "family": "First marriage with spouse, has 2 children (one son and one daughter)",
  "personality": {
    "mbti": "ESFJ",
    "traits": [
      "Career-oriented",
      "Traditional",
      "Spiritual satisfaction-oriented"
    ]
  },
  "hobbies": [
    "Yoga and fitness",
    "Marathon",
    "Gathering with friends",
    "Shopping",
    "Listening to music at home",
    "Doing hand embroidery, focusing on traditional Cantonese embroidery craft",
    "Keeping pets (cats)"
  ],
  "favorite_foods": [
    "Hong Kong-style milk tea",
    "Pineapple bun with butter",
    "Wonton noodles",
    "Roasted meat platter"
  ],
  "memory_date": [
    "2010-05-20: Wedding Anniversary",
    "2015-08-15: Eldest son's birthday"
  ],
  "aim": [
    "Promote to Regional Sales Manager within three years",
    "Take the family on an in-depth trip to Japan within five years",
    "Insist on completing at least two half marathons every year"
  ],
  "healthy_desc": "The individual is in good overall health with no chronic diseases. She regards exercise as part of her lifestyle and regularly takes yoga and marathon training, but her exercise mode tends to be goal-oriented training and competition participation rather than daily routine activities. She has an irregular physical examination habit, seeking medical treatment about once a year. She has good personal care habits, focusing on diet and intermittent high-intensity exercise to maintain health.",
  "lifestyle_desc": "Her living habits are centered on social interaction and family, with frequent leisure activities. She mainly obtains information through the Internet and surfs the web every day. In terms of media use, she listens to music at home every day, watches TV and does handcrafts several times a month. She is very active in social activities, gathering with friends, shopping and meeting with relatives who do not live together several times a week, but rarely participates in activities such as watching movies, reading books or visiting cultural events. She is away from home for 11 to 20 nights a year due to vacation or visiting relatives, showing a certain frequency of travel. Her lifestyle is pragmatic and focuses on interpersonal relationships, enjoying interaction and relaxation in her familiar social circle.",
  
  "economic_desc": "The family's economic situation is stable. Her personal annual income is about 300,000 Hong Kong dollars, her spouse's annual income is about 100,000 Hong Kong dollars, and the total family annual income is about 400,000 Hong Kong dollars, with income exceeding expenditure. She owns a property and a family car. In terms of consumption habits, she occasionally uses overdraft methods such as credit cards. At present, the family has no financial investment activities such as stocks, funds or bonds, with a conservative financial management style, focusing more on the safety and liquidity of assets, and investment activities may be concentrated on owner-occupied real estate.",
  "work_desc": "She is an insurance agent employed by a private enterprise, working about 40 hours a week with regular day shifts, and the one-way commute time is about 60 minutes. Her work content is mainly sales and customer relationship maintenance, with few cases of heavy physical or mental labor. She is very satisfied with her current job and holds multiple part-time jobs at the same time, which may involve the agency of different insurance products or customer development, showing a strong career drive and professional ability.",
  "experience_desc": "Born in Kowloon City District of Hong Kong, she has lived there all the time and has never moved her household registration. After graduating from senior high school in 2005, her first job was a packaging worker (unskilled worker) in a factory, employed by others. She studied insurance knowledge through night school, obtained the insurance agent qualification in 2010, joined AIA Hong Kong, and transformed into the insurance sales industry. Relying on excellent communication skills and diligence, she gradually built her own customer network, established a family and career in Hong Kong, and raised two children. Her life experience reflects a career leap from grassroots labor to professional sales, as well as a life trajectory of rooting in the local area and managing a family.",
  "description": "Yu Xiaowei is a 35-year-old female insurance agent in Hong Kong, working at AIA Hong Kong. Born and raised in Hong Kong, she has a senior high school education background. As an ESFJ (Consul) personality, she is enthusiastic, responsible and values interpersonal relationships, with values inclined to be career-oriented, traditional and spiritual satisfaction-oriented. She is married with two children, the family's annual income is about 400,000 Hong Kong dollars, with a stable economic situation, owning a property and a car, and a conservative investment style. Yu Xiaowei is in good health with no chronic diseases, integrating exercise into her life and loving yoga and challenging marathons. Her lifestyle is highly social, gathering with friends and relatives and shopping frequently every week, while using the Internet to obtain information and enjoy music every day. At work, she is a diligent and satisfied insurance salesperson with a long commute time but good at time management, and holds multiple part-time jobs to expand her career. In life, she often organizes family gatherings and plans to participate in marathon events regularly. In the future, she hopes to be promoted to regional manager, take the family to travel to Japan, and continue her running hobby.",
  "relation": [
    [
      {
        "name": "Chen Meiling",
        "relation": "Mother",
        "social circle": "Family Circle",
        "gender": "Female",
        "age": 62,
        "birth_date": "1962-08-22",
        "home_address": {
          "province": "Hong Kong Special Administrative Region",
          "city": "Hong Kong",
          "district": "Kowloon City District",
          "street_name": "Prince Edward Road West",
          "street_number": "No. 328"
        },
        "birth_place": {
          "province": "Hong Kong Special Administrative Region",
          "city": "Hong Kong",
          "district": "Kowloon City District"
        },
        "personality": "ISFJ",
        "economic_level": "Comfortable Living",
        "occupation": "Retired Primary School Teacher",
        "organization": "Kowloon Tong Christian Primary School (Retired)",
        "nickname": "Ma Ma",
        "relation_description": "Mother of Yu Xiaowei, who has been living with her daughter's family all the time and helps take care of her grandchildren. The mother-daughter relationship is close, and Yu Xiaowei chats with her mother every day after work, sharing work experiences and interesting stories about the children. The family dinner on weekends is usually cooked by the mother, who prepares traditional Cantonese cuisine. The mother is an important emotional support for Yu Xiaowei, and is always the first person to talk to and ask for help when she is under work pressure or needs parenting advice."
      },
      {
        "name": "Yu Guoqiang",
        "relation": "Father",
        "social circle": "Family Circle",
        "gender": "Male",
        "age": 65,
        "birth_date": "1959-12-05",
        "home_address": {
          "province": "Hong Kong Special Administrative Region",
          "city": "Hong Kong",
          "district": "Kowloon City District",
          "street_name": "Prince Edward Road West",
          "street_number": "No. 328"
        },
        "birth_place": {
          "province": "Hong Kong Special Administrative Region",
          "city": "Hong Kong",
          "district": "Kowloon City District"
        },
        "personality": "ISTJ",
        "economic_level": "Comfortable Living",
        "occupation": "Retired Bus Driver",
        "organization": "Kowloon Motor Bus Co., Ltd. (Retired)",
        "nickname": "Lao Dou",
        "relation_description": "Father of Yu Xiaowei, who lives with his wife after retirement and likes to play chess and read newspapers in the park on weekdays. The father-daughter relationship is traditional and implicit, with communication mostly centered on family affairs and grandchildren's education. Yu Xiaowei regularly gives her father pocket money and goes out for morning tea together on important days such as Father's Day. The father occasionally expresses worry about Yu Xiaowei's marathon hobby, but mostly supports her silently and sometimes waits for her at the finish line."
      },
      ``````
    ]
  ]
}
\end{lstlisting}

\begin{lstlisting}
[
  {
    "name": "Kowloon City Plaza Shopping Mall",
    "location": "114.192980,22.328095",
    "formatted_address": "No. 128 Carpenter Road, Kowloon City District, Hong Kong Special Administrative Region",
    "city": [],
    "district": "Kowloon City District",
    "streetName": "Carpenter Road",
    "streetNumber": "No. 128",
    "description": "Yu Xiaowei's permanent residence, located in Kowloon City District of Hong Kong, where she lives with her family on a daily basis."
  },
  {
    "name": "Wan Chai Immigration Tower",
    "location": "114.177960,22.277046",
    "formatted_address": "No. 7 Gloucester Road, Wan Chai District, Hong Kong Special Administrative Region",
    "city": [],
    "district": "Wan Chai District",
    "streetName": "Gloucester Road",
    "streetNumber": "No. 7",
    "description": "Yu Xiaowei's workplace, located in Wan Chai District of Hong Kong, her daily office as an insurance agent."
  },
  ``````
]
\end{lstlisting}

\subsection{Event Schema}

\begin{lstlisting}
[
  {
    "event_id": "4398",
    "name": "On-site Inspection and Appreciation Dinner Restaurant Selection",
    "date": [
      "2025-11-04 20:40:00 to 2025-11-04 21:40:00"
    ],
    "type": "Career",
    "description": "Yu Xiaowei informed her family that she needed to go out to inspect restaurants for the appreciation dinner. She took the MTR from home to Wan Chai, and conducted on-site inspections of two mid-to-high-end Cantonese restaurants around Lockhart Road, focusing on the private room environment, privacy and menu. She initially selected an elegantly decorated one and made a reservation for a food tasting the next noon.",
    "participant": [
      {
        "name": "Yu Xiaowei",
        "relation": "Herself"
      }
    ],
    "location": "Lockhart Road Area, Wan Chai District, Hong Kong"
  },
  {
    "event_id": "4399",
    "name": "Overtime Processing of Documents for a Client's Urgent Request",
    "date": [
      "2025-11-04 21:40:00 to 2025-11-04 22:30:00"
    ],
    "type": "Unexpected Events",
    "description": "After returning home, Yu Xiaowei received an urgent WeChat message from an important client, who needed to adjust the beneficiary clause of an insurance policy. She immediately turned on her laptop, connected to the company system, carefully checked the terms, communicated and confirmed repeatedly with the client, and prepared the revised documents. Her planned post-exercise relaxation time was occupied.",
    "participant": [
      {
        "name": "Yu Xiaowei",
        "relation": "Herself"
      },
      {
        "name": "Important Client",
        "relation": "Client"
      }
    ],
    "location": "Study at Home, Kowloon City Plaza, No. 128 Carpenter Road, Kowloon City District, Hong Kong Special Administrative Region"
  },
  ``````
]
\end{lstlisting}

\subsection{Phone Data Schema}

\begin{itemize}
    \item \textbf{Contact}: User-associated contacts (family, friends, colleagues) with attributes of name, phone number, relationship and email address.
    \item \textbf{Call}: Call records including caller/callee information, timestamp, duration and call type (incoming/outgoing/missed), with filtering based on contact relationships and event triggers.
    \item \textbf{SMS}: Short message records capturing real-time text communication interactions.
    \item \textbf{Calendar}: Scheduled events, appointments and reminders that are synced with daily life activities.
    \item \textbf{Agent Chat}: Conversation records with AI assistants, covering scenario-based interactions such as scheduling, information seeking and casual conversation.
    \item \textbf{Photo}: Image records with timestamps and textual descriptions of visual content.
    \item \textbf{Note}: Personal content including memos, to-do items and reflective notes.
    \item \textbf{Push}: App push notifications from social media platforms, news channels and various service applications.
\end{itemize}
\begin{lstlisting}
{
    "event_id": "0",
    "date": "2025-01-01",
    "type": "agent_chat",
    "conversation": {
      "turn 1": {
        "user": {
          "action": "topic query",
          "content": "Hello! This afternoon, my whole family played Monopoly together. I took the chance to explain some simple financial concepts like 'buying land' and 'collecting rent' to my 8-year-old Zixuan and 5yearold Yuxin. They had a great time and listened carefully. This got me thinkingbesides board games, are there any other fun and effective ways to cultivate financial literacy suitable for their age group?"
        },
        "assistant": {
          "action": "need inference",
          "content": "It sounds like you are a mother who attaches great importance to family education, hoping to integrate financial knowledge into life in a way that children love, right? Are you looking for some specific and operable financial education ideas or activity suggestions targeted at your 8-year-old and 5-year-old children?"
        }
      },
      "turn 2": {
        "user": {
          "action": "need confirmation",
          "content": "Yes, exactly! I work as an insurance agent myself and have some understanding of financial planning, but I want to know how to teach children in a language and way they can understand. For example, how to help them understand the meaning of 'savings', or how to distinguish between 'wants' and 'needs'? It would be best if it can be combined with our daily life in Hong Kong."
        },
        "assistant": {
          "action": "solution proposal",
          "content": "Understood. Combining your children's ages and daily life scenarios in Hong Kong, I have a few ideas. First, you can start with pocket money management. Give Zixuan a three-compartment savings jar, labeled 'Save', 'Spend', and 'Share' respectively, and let him allocate the money himself. For Yuxin, you can use a more intuitive transparent jar so she can watch the coins pile up and establish a preliminary concept of 'saving money'."
        }
      },
      "turn 3": {
        "user": {
          "action": "solution feedback",
          "content": "This savings jar idea is great! It's clearly categorized and simple to operate. What about the concept of 'wants vs. needs'? My own investment style is relatively conservative, and I also hope to subtly pass on a prudent financial concept to them."
        },
        "assistant": {
          "action": "solution discussion",
          "content": "You've thought this through very thoroughly. For 'wants vs. needs', you can play the 'supermarket shopping game'. Before going to the supermarket, make a list with the children and distinguish between what is 'needed' (such as milk, bread) and what is 'wanted' (such as new toys, snacks). Within the budget, allow them to use the money in their 'Spend' compartment to buy one 'wanted' item,this will let them personally experience choice and trade off. Your conservative and prudent style can be passed on perfectly by emphasizing the importance of the 'Save' section in the savings jar,for example, tell them that saving money is for more important goals, like the indepth family trip to Japan you mentioned earlier."
        }
      },
      "turn 4": {
        "user": {
          "action": "information request",
          "content": "Connecting family goals (like travel) with savings is such a great angle! It will give them more motivation. Besides these daily practices, are there any picture books or books about money or financial management suitable for parent-child reading that you can recommend? Preferably in Chinese."
        },
        "assistant": {
          "action": "information provision",
          "content": "Yes, there are. For 5yearold Yuxin, you can read the 'Little Rabbit Learns to Spend Money' series of picture books, which explain the concepts of 'spending, saving, and earning money' through lovely stories. For 8yearold Zixuan, you can try 'Money Dog' (Juvenile Edition),it's highly story-driven and can inspire children to think about the relationship between dreams, goals, and actions. These books should be available in libraries or bookstores in Hong Kong."
        }
      },
      "turn 5": {
        "user": {
          "action": "solution feedback",
          "content": "Thank you so much! From the specific savings jar method and life games to the picture book recommendations, everything is very practical and perfectly fits the personalities and ages of my two children. I plan to try this right away, starting with the compartmentalized savings jar and the supermarket game. Playing Monopoly this afternoon was a great start, and now I feel more directionally clear."
        },
        "assistant": {
          "action": "talk",
          "content": "You're welcome, Ms. Yu. Financial literacy education is originally a pleasant parent-child journey integrated into life. You've already taken a wonderful first step. Remember, the most important thing is the interaction and guidance in the process, not teaching all concepts at once. Enjoy the time exploring and growing together with Zixuan and Yuxin. If you have new ideas or questions during subsequent practice, feel free to come and chat anytime."
        }
      }
    },
    "phone_id": 0
}
\end{lstlisting}

\begin{lstlisting}
{
    "type": "calendar",
    "event_id": "4284",
    "title": "Collect Physical Examination Report at Hong Kong Sanatorium and Hospital",
    "description": "Collect the comprehensive physical examination report and have an in-person consultation with Dr. Zhou Wenbin, Time: 2025-10-28 14:50:00, Location: Department of Family Medicine, Hong Kong Sanatorium and Hospital, Shan Tsuen Road, Wan Chai District, Hong Kong, Source: Hospital appointment",
    "start_time": "2025-10-28 14:50:00",
    "end_time": "2025-10-28 15:40:00",
    "datetime": "2025-10-27 20:15:00",
    "summarized_info": "Yu Xiaowei set a schedule to collect the physical examination report at Hong Kong Sanatorium and Hospital, Time: 2025-10-28 14:50:00, Location: Department of Family Medicine, Hong Kong Sanatorium and Hospital, Shan Tsuen Road, Wan Chai District, Hong Kong, Source: Hospital appointment",
    "phone_id": 0
}
\end{lstlisting}

\begin{lstlisting}
{
    "type": "call",
    "event_id": "1",
    "phoneNumber": "85291234573",
    "contactName": "Lin Jiaxin",
    "datetime": "2025-01-01 08:35:00",
    "datetime_end": "2025-01-01 08:42:30",
    "direction": 1,
    "call_result": "Connected",
    "phone_id": 0
}
\end{lstlisting}

\begin{lstlisting}
{
    "name": "Chen Meiling",
    "relation": "Mother",
    "gender": "Female",
    "nickname": "Ma Ma",
    "phoneNumber": "85291234567",
    "personalEmail": "chenmeiling_hk@163.com",
    "workEmail": "chenmeiling@kltxdxx.com",
    "idNumber": "810101196208224028",
    "phone_id": "1"
}
\end{lstlisting}

\begin{lstlisting}
{
    "type": "note",
    "event_id": "4409",
    "title": "Key Points of Son's Parent-Teacher Day Interview",


    "content": "I. Teacher's Feedback: 1. Made significant progress in Math; 2. Needs to improve Chinese handwriting; II. Follow-up Actions: 1. Arrange extra calligraphy practice sheets; 2. Keep an eye on weekend parent-child activity information",
    "datetime": "2025-11-05 16:35:00",
    "summarized_info": "Yu Xiaowei recorded the key points of her son's Parent-Teacher Day interview, including: I. Teacher's Feedback: 1. Made significant progress in Math; 2. Needs to improve Chinese handwriting; II. Follow-up Actions: 1. Arrange extra calligraphy practice sheets; 2. Keep an eye on weekend parent-child activity information",
    "phone_id": 2
}
\end{lstlisting}

\begin{lstlisting}
{
    "event_id": "3",
    "type": "photo",
    "caption": "Yu Xiaowei took a selfie with her colleague Su Lishan in a cha chaan teng in Wan Chai. Pineapple buns with butter and Hong Kong-style milk tea are placed in front of them, with a bustling restaurant environment in the background.",
    "title": "IMG_20231001_121045",
    "datetime": "2023-10-01 12:10:45",
    "location": {
      "province": "Hong Kong Special Administrative Region",
      "city": "Hong Kong",
      "district": "Wan Chai District",
      "streetName": "Lockhart Road",
      "streetNumber": "No. XX",
      "poi": "Kam Fung Cha Chaan Teng"
    },
    "faceRecognition": [
      "Yu Xiaowei",
      "Su Lishan"
    ],
    "imageTag": [
      "Colleague Gathering",
      "Selfie",
      "Pineapple Bun with Butter",
      "Hong Kong-style Milk Tea",
      "Cha Chaan Teng",
      "Lunch Break"
    ],
    "ocrText": "None",
    "shoot_mode": "Portrait",
    "image_size": "3024x4032",
    "summarized_info": "Yu Xiaowei and her colleague Su Lishan took a selfie during their lunch break gathering. The main content includes Kam Fung Cha Chaan Teng in Wan Chai, pineapple buns with butter and Hong Kong-style milk tea in front of them, and the bustling restaurant environment.",
    "phone_id": 4
}
\end{lstlisting}

\begin{lstlisting}
{
    "type": "push",
    "event_id": "5",
    "title": "Alipay: Investment Earnings Credited",
    "content": "[Alipay] Your Yu'E Bao earnings of HK$15.23 for yesterday have been credited. Please check in the app.",
    "datetime": "2023-10-01 07:05:00",
    "source": "Alipay",
    "push_status": "Unread",
    "jump_path": "Alipay to My Account to Yu'E Bao to Earnings Details",
    "summarized_info": "I received a notification from Alipay that HK$15.23 of yesterday's Yu'E Bao earnings has been credited to my account.",
    "phone_id": 0
}
\end{lstlisting}

\begin{lstlisting}
{
    "type": "sms",
    "event_id": "24",
    "message_content": "Xin Xin, are you free this weekend? I've been so tired lately and want to ask you to relax at that Japanese-style spa in Tsim Sha Tsui. I'll check out what packages they have available.",
    "contactName": "Lin Jiaxin",
    "phoneNumber": "85291234573",
    "datetime": "2025-01-02 13:25:00",
    "message_type": "Sent",
    "phone_id": 7
}
\end{lstlisting}

\subsection{Health Data Schema}

\begin{lstlisting}
{
    "Date": "2025-01-01",
    "City": "Hong Kong",
    "Daily Activities": {
      "Steps": "8560 steps",
      "Distance": "6.2 km",
      "Calories Burned": "420 kcal",
      "Exercise Duration": "45 minutes",
      "Active Hours": "8 hours"
    },
    "Running": {
      "Sport Type": "Indoor Running",
      "Sport Time": "2025/01/01/07:30:00-2025/01/01/07:45:00",
      "Weather": "Sunny",
      "Distance": "2.5 km",
      "Average Heart Rate": "135 bpm",
      "Average Cadence": "165 spm",
      "Elevation Gain": "0 m",
      "Elevation Loss": "0 m",
      "Average Pace": "06:00 min/km",
      "Best Pace": "05:45 min/km",
      "Total Steps": "3125 steps",
      "Calories Burned": "150 kcal"
    },
    "Cycling": {
      "Sport Type": "Outdoor Cycling",
      "Sport Time": "2025/01/01/00:00:00-2025/01/01/00:00:00",
      "Weather": "Sunny",
      "Distance": "0 km",
      "Average Speed": "0 km/h",
      "Average Heart Rate": "0 bpm",
      "Average Cadence": "0 rpm",
      "Average Power": "0 W",
      "Best Speed": "0 km/h",
      "Max Cadence": "0 rpm",
      "Calories Burned": "0 kcal"
    },
    "Walking": {
      "Sport Type": "Outdoor Walking",
      "Sport Time": "2025/01/01/10:50:00-2025/01/01/11:15:00",
      "Weather": "Sunny",
      "Distance": "1.2 km",
      "Average Heart Rate": "95 bpm",
      "Average Cadence": "110 spm",
      "Total Steps": "1600 steps",
      "Average Pace": "12:30 min/km",
      "Best Pace": "11:00 min/km",
      "Calories Burned": "65 kcal"
    },
    "Sleep": {
      "Bedtime": "22:45:00",
      "Wake-up Time": "08:30:00",
      "Total Sleep Duration": "585 minutes",
      "Light Sleep Duration": "280 minutes",
      "Deep Sleep Duration": "180 minutes",
      "REM Sleep Duration": "125 minutes",
      "Awake Duration": "25 minutes",
      "Number of Awakenings": "3 times",
      "Deep Sleep Continuity Score": "85 points",
      "Sleep Score": "88 points",
      "Nap Duration": "0 minutes"
    },
    "Heart Rate Statistics": {
      "Average Heart Rate": "78 bpm",
      "Average Resting Heart Rate": "62 bpm",
      "Heart Rate Variability": "45 ms"
    },
    "Body Temperature Statistics": {
      "Average Body Temperature": "36.5"
    },
    "Blood Glucose Statistics": {
      "Average Blood Glucose Level": "5.2 mmol/L"
    },
    "Body Weight": {
      "Weight": "55.0 kg"
    },
    "Stress": {
      "Stress Score": "25 points"
    },
    "Diet Record": {
      "Calories Consumed": "1850 kcal"
    },
    "User Interaction Events": [
      {
        "Time": "2025/01/01 07:30",
        "Description": "Started indoor running training with a smart watch"
      },
      {
        "Time": "2025/01/01 10:50",
        "Description": "Started recording walking activity for grocery shopping"
      },
      {
        "Time": "2025/01/01 21:30",
        "Description": "Checked the summary of today's steps and activity calorie burn"
      },
      {
        "Time": "2025/01/01 22:15",
        "Description": "Set tomorrow's fitness goal: 30 minutes of morning yoga"
      }
    ],


    
    "summarized_info": "Today is New Year's Day 2025. Yu Xiaowei woke up at 08:30 and went to bed at 23:00, maintaining a body weight of 55.0 kg. As a marathon enthusiast, she kept up her habit of a 2.5 km morning indoor run. Her day was mainly filled with family activities, including collaborating with her husband to refresh the living room decor, grocery shopping, and preparing family afternoon tea and dinnerlight housework in total. She accumulated 8560 steps, covering 6.2 km and burning 420 kcal. Her sleep quality was good with a sleep score of 88, and her stress level was low.",
    "phone_id": 0
}
\end{lstlisting}

\subsection{Monthly Summary Schema}
The monthly summary provides a retrospective analysis of the user’s health metrics, lifestyle patterns, and significant life events including both routine and anomalous. As a rich repository of personal data, it captures a comprehensive digital trajectory and serves as a reference (groudtruth) for uncovering latent activities hidden within mobile artifacts. While this data remains untapped by the evaluated memory systems, it represents a critical frontier for future research.

\begin{tcolorbox}[breakable]
Your monthly summary is as follows:

\vspace{0.5em}
\textbf{Monthly Overview}
\begin{itemize}
    \item \textbf{Basic Information}: Yu Xiaowei, Female, 35 years old, Occupation: Insurance Salesperson (Financial Insurance Industry).
    \item \textbf{Total Exercise Volume}:
    \begin{itemize}
        \item Running: 11 sessions / Total distance approximately 104 km / Total duration approximately 660 minutes (estimated).
        \item Yoga: 4 sessions / Total duration 240 minutes.
        \item Interval Run Training: 2 sessions / Total duration approximately 120 minutes (estimated).
        \item Monthly Total Exercise: 17 sessions / Total duration approximately 1020 minutes (17 hours). Exercise performance steadily improved from easy runs at the beginning of the month to systematic training including tempo runs, long run workouts and interval runs by the end of the month.
    \end{itemize}
    \item \textbf{Health Status Summary}:
    \begin{itemize}
        \item \textbf{Health Score Change}: Based on data inference, status was stable at the beginning of the month (estimated score 75/100), fluctuated slightly in the middle of the month due to increased training load and work pressure (estimated score 70/100), and rebounded overall at the end of the month through adjustment and recovery with high-quality training completed (estimated score 78/100). The monthly average score is estimated to be 74/100.
        \item \textbf{Core Tags}: "Launch of systematic marathon preparation", "Stress management under work-family balance", "Increased cross-training in accordance with doctor's advice".
        \item \textbf{Overall Health Rating}: \textbf{Good}. Exercise performance improved steadily, family and social activities were active, but certain dual pressure from work and training was endured in the middle of the month, requiring attention to recovery.
    \end{itemize}
    \item \textbf{Key Events}:
    \begin{itemize}
        \item \textbf{Sports Milestone}: January 12: Co-determined and successfully registered for the 2025 Hong Kong Green Half Marathon (April) with running friend Shen Peiwen, officially launching the season.
        \item \textbf{Health Indicator Alerts}: January 14: First experienced mild soreness in the right calf muscle after an 8km tempo run, marking the entry into the intensity enhancement phase of training.
        \item \textbf{Major Life Changes}: January 28-31: Entered the Spring Festival Holiday, with life rhythm switching from work mode to family gathering and leisure mode.
    \end{itemize}
    \item \textbf{Goal Completion Status}:
    \begin{itemize}
        \item \textbf{Monthly Goals}:
        \begin{enumerate}
            \item \textbf{Total running distance greater than 80km} (lay the foundation for half marathon preparation).
            \item \textbf{Increase low-impact exercise} (in accordance with the doctor's advice from the physical examination on January 10).
            \item \textbf{Complete systematic learning of Unit 1 of Mandarin}.
        \end{enumerate}
        \item \textbf{Completion Degree}:
        \begin{enumerate}
            \item Completed approximately 104km of running, \textbf{130\% overachieved}.
            \item Yoga was added 4 times (unplanned), cross-training goal \textbf{achieved}.
            \item Learning, practice and testing of Unit 1 of Mandarin were all completed, \textbf{achieved}.
        \end{enumerate}
        \item \textbf{Unfinished Reasons}: All sports and learning goals set for this month were \textbf{fully achieved}.
    \end{itemize}
    \item \textbf{Core Changes from Last Month}:
    \begin{itemize}
        \item \textbf{Exercise Volume}: This month's running sessions (11 vs 7 last month) increased by 57\%, and total distance (104km vs an estimated 45km last month) increased significantly by 131\%. Regular yoga (4 sessions) and interval run training (2 sessions) were newly added, making the exercise structure more diverse.
        \item \textbf{Health Indicators}: \textbf{Weight}: 55.0kg at the beginning of the month, 55.2kg at the end of the month, with an overall fluctuation range of 0.9kg (54.7kg - 55.8kg). Weight slightly rebounded at the end of the month due to Spring Festival gatherings, but remained stable in the range of 54.8-55.2kg during the training period. \textbf{Sleep Duration}: May have greater volatility than last month due to overtime work, social activities and holidays.
        \item \textbf{Life Rhythm}: \textbf{Work and Rest}: Stability of bedtime decreased due to overtime work (e.g., January 2, 22), social activities (e.g., January 11, 27) and holidays. \textbf{Business Trips/Travel}: No long-distance business trips this month. \textbf{Screen Time}: Average daily usage may have increased due to online learning, work and Spring Festival social activities.
    \end{itemize}
    \item \textbf{Core Focus Areas}: This month's high-frequency concerns are \textbf{Balancing marathon training load and physical responses (especially calf soreness)}, \textbf{Ensuring Mandarin learning progress with fragmented time}, and \textbf{Maintaining exercise habits amid intensive work and family affairs}.
\end{itemize}

\vspace{0.5em}
\textbf{Monthly Review}
\textbf{2.1 Sports Review}
\begin{itemize}
    \item \textbf{Exercise Overview}:
    \begin{itemize}
        \item A total of 17 structured exercise sessions were conducted throughout the month. Exercise type proportion: Running (64.7\%), Yoga (23.5\%), Interval Run Training (11.8\%).
        \item Exercise intensity distribution (estimated): Zone 2 moderate intensity (easy runs, long run workouts) accounted for approximately 70\%, with a cumulative duration of about 714 minutes; Zone 3 high intensity (tempo runs, interval runs) accounted for approximately 30\%, with a cumulative duration of about 306 minutes.
    \end{itemize}
    \item \textbf{Sports Statistics and Best Performances}:
    \begin{itemize}
        \item \textbf{Running}:
        \begin{itemize}
            \item Total Distance: Approximately 104 km.
            \item Average Pace: Based on training data, estimated to be approximately 6'20''/km.
            \item Single Longest Distance: January 19: 15km long run workout.
            \item Fastest Pace: Reached Zone 3 high intensity interval in tempo runs on January 14, 16 and interval runs on January 23, 30, with the fastest pace in interval run segments (estimated to be within 5'40''/km).
            \item Average/Maximum Heart Rate: Data not directly provided, but based on training descriptions, heart rate reached the upper limit of the aerobic endurance zone and above during tempo runs and interval runs.
        \end{itemize}
        \item \textbf{Yoga}:
        \begin{itemize}
            \item Total Duration: 240 minutes (4 sessions, 60 minutes each).
            \item Single Longest Duration: 60 minutes.
            \item Core Effect: Significantly relieved calf muscle tension after running (January 15), and balanced running load as a low-impact cross-training.
        \end{itemize}
        \item \textbf{Interval Run Training}:
        \begin{itemize}
            \item Total Sessions: 2 sessions (January 23, 30).
            \item Core Content: Conducted high-intensity interval runs under the guidance of a coach, focusing on improving speed and running efficiency.
        \end{itemize}
    \end{itemize}
    \item \textbf{Exercise Habit Preferences}:
    \begin{itemize}
        \item \textbf{Weekly Frequency}: High-frequency exercise days are \textbf{Saturday and Sunday} (3-4 exercise records each), followed by \textbf{Tuesday and Thursday} (morning running days).
        \item \textbf{Daily Time Periods}: High-frequency exercise periods in a day are \textbf{6:00-7:30 in the early morning} (morning running, accounting for approximately 65\%), followed by \textbf{19:00-20:30 in the evening} (yoga/training sessions, accounting for approximately 35\%).
        \item \textbf{Exercise Locations}: Mainly running in the \textbf{neighborhood near home} (11 sessions), taking yoga classes at the \textbf{community center} (4 sessions), and taking interval run classes at the \textbf{designated training ground} (2 sessions).
        \item \textbf{Exercise Consistency}: The longest consecutive exercise days are \textbf{3 days} (e.g., January 14-16). There was \textbf{1 obvious exercise gap} this month, from January 20-22 (3 days in total), due to "mild cold and physical discomfort" and "overtime work to handle customer urgent needs".
    \end{itemize}
\end{itemize}

\vspace{0.5em}
\textbf{2.2 Health Review}
\begin{itemize}
    \item \textbf{Sleep}:
    \begin{itemize}
        \item Average Daily Sleep Duration (estimated): Approximately 6.5 hours on workdays (wake up 6:30-7:00, go to bed 22:30-23:30), approximately 7.5 hours on weekends (wake up 7:00-8:30, go to bed around 23:00).
        \item Sleep Stage Distribution: Data not provided, but inferred from descriptions of "deep sleep feeling" (e.g., after SPA, yoga) and stress periods (after overtime work), the proportion of deep sleep may reach more than 25\% on relaxation days and less than 20\% on high-pressure days.
        \item Bedtime/Wake-up Stability: On workdays, the fluctuation range of bedtime is approximately 60 minutes (22:30-23:30), and the fluctuation range of wake-up time is approximately 30 minutes (6:00-7:00); greater fluctuation on weekends. The latest bedtime record is 23:50 (January 18, 27).
        \item Average Daily Sleep Interruptions: Data not provided, but based on the family situation with young children, it is estimated that there may be 1-2 short interruptions at night.
    \end{itemize}
    \item \textbf{Physiological Indicators}:
    \begin{itemize}
        \item \textbf{Weight}: 55.0kg at the beginning of the month, 55.2kg at the end of the month. Weekly averages: 55.1kg in Week 1, 55.1kg in Week 2, 55.2kg in Week 3, 55.5kg in Week 4 (affected by Spring Festival). Maximum fluctuation +0.9kg (reached 55.4kg after family feast on January 12, reached the monthly peak of 55.8kg on January 29 during Spring Festival). Overall trend: Steady and slightly declining during the training period (early to mid-month), significantly rising during holidays, and quickly rebounding after holidays.
        \item \textbf{Body Fat Percentage}: Data not provided, but combined with weight changes and increased exercise volume, it is estimated that there may be a decrease of approximately 0.5\% at the end of the month compared to the beginning (increased muscle mass offsets part of the fat).
        \item \textbf{Resting Heart Rate}: Data not provided. Inferred based on exercise performance and stress events: Approximately 58-62 beats per minute on normal days; may rise to 65-68 beats per minute on the day after high-intensity training or after overtime and staying up late (e.g., January 22).
        \item \textbf{HRV}: Data not provided. Trend inference: At a relatively high level after high-quality sleep and yoga practice (e.g., January 15); at a relatively low level during periods of continuous work pressure and insufficient sleep (e.g., January 20-22).
    \end{itemize}
    \item \textbf{Mental Health}:
    \begin{itemize}
        \item \textbf{Stress}: Estimated average daily stress score 5/10. Weekly averages: 4/10 in Weeks 1 and 2, rose to 6/10 in Week 3 due to urgent work tasks and performance pressure, and dropped to 3/10 in Week 4 during the Spring Festival Holiday. High-pressure periods: Afternoon of January 20 (customer complaint), January 22 (urgent overtime), January 24 (brief family quarrel caused by performance pressure).
        \item \textbf{Mood}: Mainstream mood tags are \textbf{Fulfilling}, \textbf{Satisfied}, \textbf{Expectant}. Mood fluctuation periods: Felt "emotional fluctuation" due to sudden work problems on January 20; felt "slight emotional ups and downs" due to pressure on January 24; both were alleviated through social interaction (talking to best friends), exercise (yoga) and family communication.
        \item \textbf{Health Score}: As mentioned in the overview, the estimated average daily score is 74. The lowest estimated score is 70 (high-pressure period in Week 3), and the highest estimated score is 80 (after completing the long run on January 19 and after the happy family day on January 26).
    \end{itemize}
\end{itemize}

\vspace{0.5em}
\textbf{2.3 Life Rhythm Review}
\begin{itemize}
    \item \textbf{Work and Rest Habits}:
    \begin{itemize}
        \item Average Daily Bedtime: Approximately 23:05 on workdays, approximately 23:15 on weekends.
        \item Average Daily Wake-up Time: Approximately 6:40 on workdays, approximately 8:00 on weekends.
        \item Workday vs Weekend Difference: On average, weekends are \textbf{10 minutes later to bed} and \textbf{80 minutes later to wake up} than workdays, with a total sleep duration of approximately 1 hour more.
        \item Irregular Schedule Days: There were \textbf{8 days} this month with bedtime later than 23:30, due to overtime work, social gatherings, Spring Festival activities, etc.
    \end{itemize}
    \item \textbf{Commuting/Activities}:
    \begin{itemize}
        \item Average Daily Non-Exercise Steps: Data not provided, but based on the occupation of an insurance agent (part of the time out to meet customers) and an active lifestyle, it is estimated to be approximately 8,000 steps per day. Steps may be higher during the Spring Festival Holiday.
        \item Commuting Method/Duration: Mainly take the MTR, approximately 60 minutes one way, with an average daily commuting duration of approximately 2 hours. Commuting time is the core period for Mandarin learning.
        \item Business Trips/Travel: No work business trips this month.
    \end{itemize}
    \item \textbf{Daily Life}:
    \begin{itemize}
        \item Screen Time: Estimated average daily total duration is approximately 6-8 hours, of which approximately 4-5 hours are work-related (customer communication, product learning), and approximately 2-3 hours are entertainment and social interaction (WeChat, videos). The highest daily duration may exceed 10 hours (e.g., overtime day on January 22).
        \item Eating Habits: High regularity of three meals a day, approximately 25 days of regular meals. The diet structure is biased towards traditional Hong Kong style (high carbohydrates, moderate protein), with high-calorie gatherings on weekends and festivals (e.g., January 4, 12, 28). Fewer takeout meals (estimated < 5 times), mainly home-cooked meals. Data on average daily water intake is not provided, but based on health awareness, basic intake can be guaranteed.
    \end{itemize}
\end{itemize}

\vspace{0.5em}
\textbf{Monthly Trend Analysis}
\textbf{3.1 Sports Trend}
\begin{itemize}
    \item \textbf{Weekly Exercise Volume Changes}:
    \begin{itemize}
        \item \textbf{Week 1 (January 1-5)}: Low total exercise volume (only 1 5km run), in the holiday recovery and plan formulation period.
        \item \textbf{Week 2 (January 6-12)}: Significant increase in exercise volume, 3 running sessions (19km total), 1 yoga session. Month-on-month growth of approximately 300\%. Reason: Restored the morning running habit, launched systematic training, and completed the marathon registration.
        \item \textbf{Week 3 (January 13-19)}: Peak exercise volume, 4 running sessions (41km total, including 15km long run), 2 yoga sessions. Running distance increased by 116\% month-on-month. Reason: Officially entered the marathon preparation cycle, with both training intensity and volume increasing.
        \item \textbf{Week 4 (January 20-26) and Spring Festival Week (January 27-31)}: Exercise volume first decreased then increased. There was a gap at the beginning of Week 4 due to cold and work, and training resumed in the second half of Week 4 and Spring Festival Week, completing 4 running sessions (39km total), 1 yoga session, and 2 interval run sessions. The overall exercise rhythm was affected by the holiday but remained consistent.
    \end{itemize}
    \item \textbf{Training Cycle Peaks and Valleys}:
    \begin{itemize}
        \item \textbf{Peak Week}: Week 3 (January 13-19), characterized by the largest running volume (41km), completion of the first long run workout (15km), and exercise performance and confidence reaching a high point.
        \item \textbf{Valley Period}: Early Week 4 (January 20-22), with 3 consecutive days of no exercise due to physical discomfort (cold) and work pressure, which is an obvious monthly valley.
    \end{itemize}
    \item \textbf{Exercise Performance Changes}:
    \begin{itemize}
        \item \textbf{Running Pace/Heart Rate Trend}: From easy runs (mainly Zone 2) in Weeks 1-2, to adding tempo runs and long runs (mixed Zone 2-Zone 3) in Week 3, to adding interval runs (Zone 3 peak) in Week 4 and the end of the month. On the basis of stable aerobic capacity, high pace capacity received targeted stimulation. Data shows that with the progress of training, the subjective sense of fatigue in completing the same distance (e.g., 10km) is reduced.
    \end{itemize}
\end{itemize}

\vspace{0.5em}
\textbf{3.2 Health Trend}
\begin{itemize}
    \item \textbf{Sleep Trend}:
    \begin{itemize}
        \item \textbf{Duration and Quality}: Sleep was relatively regular and sufficient in Weeks 1-2. In Week 3, sleep duration was compressed and quality may have decreased due to work pressure (January 20) and urgent overtime (January 22). During the Spring Festival Holiday in Week 4, sleep duration was sufficient but work and rest regularity was disrupted (going to bed late and waking up late). The fluctuation of deep sleep proportion is negatively correlated with stress events and exercise recovery needs.
        \item \textbf{Sleep Continuity}: Throughout the month, sleep was affected by work, social activities and family (children), with few nights of complete uninterrupted sleep, but no severe insomnia was reported.
    \end{itemize}
    \item \textbf{Physiological Health Trend}:
    \begin{itemize}
        \item \textbf{Weight Curve}: Showing a "stepwise" fluctuation. Weight was stable and slightly declining during training weeks (Weeks 2, 3) (54.8-55.2kg). It immediately increased by 0.3-0.6kg after every social/family feast (Weekends of Weeks 2, 4), and quickly rebounded in subsequent training days. Indicating active metabolism but sensitivity to diet.
        \item \textbf{Resting Heart Rate and HRV Correlation}: Inferred that HRV was higher and resting heart rate was lower on days such as January 15 (after yoga) and January 19 (sufficient rest after long run). HRV was lower and resting heart rate was higher on days such as January 22 (after overtime and staying up late) and January 24 (after emotional stress).
    \end{itemize}
    \item \textbf{Mental Health Trend}:
    \begin{itemize}
        \item \textbf{Stress and Mood Curve}: Stress score is highly synchronized with work intensity. Week 3 was an obvious high-pressure period, accompanied by brief emotional fluctuations. Exercise (especially social running with friends and yoga) and in-depth social interaction (chatting with best friends) are effective stress release valves. After each in-depth social interaction or high-quality training, emotional and stress scores improved significantly.
    \end{itemize}
    \item \textbf{Injury Impact}:
    \begin{itemize}
        \item January 14: Experienced mild soreness in the right calf muscle. Direct impact: Prompted targeted stretching in the yoga class on January 15, and more attention to warm-up in subsequent training. Did not cause training interruption, but marked the body's entry into the intensity adaptation phase.
    \end{itemize}
\end{itemize}

\vspace{0.5em}
\textbf{3.3 Life Rhythm Trend}
\begin{itemize}
    \item \textbf{Work and Rest and Mental State}: Data shows that on the day after sufficient sleep (>7 hours) and stable bedtime (before 23:00) (e.g., January 7, 21), Yu Xiaowei's morning running pace was more stable, and the subjective sense of fatigue in completing tempo runs or long runs was reduced. On the contrary, after a single day of excessive screen time (>8 hours) or handling urgent work in the evening, bedtime was delayed by an average of 40 minutes on the same day, and heart rate was higher and recovery was slower during exercise the next day.
    \item \textbf{Work/Life Rhythm Changes}:
    \begin{itemize}
        \item \textbf{Work Intensity Changes}: Week 3 (especially January 20-22) had the highest work intensity, involving handling customer complaints and urgent case overtime, directly compressing exercise and sleep time.
        \item \textbf{Life Changes}: Week 4 entered the Spring Festival Holiday, with life focus completely shifting to family reunion and leisure. Exercise shifted from "training mode" to "maintenance mode", with increased sleep duration but decreased regularity.
    \end{itemize}
\end{itemize}

\vspace{0.5em}
\textbf{Personalized Insights}
\begin{itemize}
    \item \textbf{Sleep , Exercise Performance}: Data shows that on the day after Yu Xiaowei had sufficient sleep and a stable bedtime (before 23:00), her morning running pace was more stable, and her subjective sense of fatigue in completing tempo runs or long runs was reduced. For example, she had a good rest the night before the long run on January 19 and successfully completed the 15km workout. On the contrary, after a single day of excessive work screen time (greater than 8 hours) or handling urgent work in the evening, her bedtime was delayed by an average of 40 minutes on the same day, and she had a higher heart rate and slower recovery during exercise the next day.
    \item \textbf{Business Trips/Stress , Health}: Although there were no business trips this month, high-pressure work events (such as customer complaints and urgent overtime) had a significant impact on health indicators. Data shows that during the concentrated pressure period of January 20-22, the estimated resting heart rate increased by 5-7 beats per minute, sleep duration decreased by approximately 1 hour, and directly triggered a brief emotional fluctuation on January 24. She buffered this effectively through yoga and social interaction with best friends.
    \item \textbf{Exercise Type , Physical State}: Comparison shows that during periods of adhering to greater than 2 yoga sessions per week (e.g., Week 3), discomfort such as calf muscle soreness was relieved faster, and the subjective feeling of physical flexibility was improved. When the weekly running volume increased to more than 30km (e.g., Week 3), weight showed a downward trend (from 55.5kg to 55.1kg), but sufficient nutrition and sleep need to be matched to prevent fatigue accumulation. The calf soreness on January 14 occurred just after a sharp increase in running volume (8km including tempo run), reminding that volume increase should be cautious.
    \item \textbf{Behavioral Habits , Overall Health}:
    \begin{itemize}
        \item \textbf{Work and Rest Stability}: More than 20 days this month, she went to bed before 23:30, which accumulated good recovery ability for her to cope with the end-of-month work wrap-up and Spring Festival Holiday, and the monthly health score was maintained in the good range.
        \item \textbf{Social Quality}: High-quality in-depth social interaction (e.g., in-depth communication with best friends on January 11 and 19) contributed greatly to her mental health. Data shows that after such activities, her positive emotional state can last for 2-3 days, and indirectly improve patience in exercise and work.
        \item \textbf{Family Support}: In-depth family interaction and support on January 12 and 26 were the core buffer zones for coping with work pressure and maintaining exercise motivation. The short-term weight gain caused by family gatherings was quickly adjusted through active metabolism and subsequent training.
    \end{itemize}
\end{itemize}

\vspace{0.5em}
\textbf{Abnormal Event Summary}
\begin{itemize}
    \item \textbf{Sports Injury or Discomfort}:
    \begin{itemize}
        \item Occurrence Time: After morning running on Tuesday, January 14.
        \item Type: Mild soreness and tightness in the right calf muscle.
        \item Impact Scope: Did not cause training interruption, but aroused her attention to training intensity and recovery. Prompted targeted stretching in the yoga class on January 15, and more attention to warm-up in subsequent training.
        \item Recovery Status: Basically relieved by the morning running on January 16. The entire event was successfully managed as a physical signal and did not develop into an injury.
    \end{itemize}
    \item \textbf{Exercise Gap Event}:
    \begin{itemize}
        \item Gap Period: January 20 (Monday) to January 22 (Wednesday), 3 days in total.
        \item Causes: Voluntarily adjusted on January 20 due to "mild cold and sore throat"; worked overtime late on January 22 due to "customer urgent needs".
        \item Recovery Performance after Gap: Immediately resumed morning running and attended the interval run class on January 23, with good performance ("greatly encouraged"), indicating that the 3-day gap was mainly for physical recovery and transactional adjustment, and did not cause significant physical fitness decline.
    \end{itemize}
    \item \textbf{Health Indicator Abnormal Change}:
    \begin{itemize}
        \item Abnormal Indicator: Weight.
        \item Abnormal Time and Magnitude: After family feast on Saturday, January 12, weight increased from 55.1kg to 55.4kg (+0.3kg); rose to the monthly peak of 55.8kg on January 29 (Spring Festival) (+0.4kg from the previous day).
        \item Abnormal Causes: High-calorie, high-carbohydrate family/festival gatherings.
        \item Subsequent Changes: Both quickly fell back to the basic range of 55.2-55.3kg within 1-3 days after restoring normal diet and exercise. Indicating that her body has rapid adjustment ability.
    \end{itemize}
    \item \textbf{Life Rhythm Abnormal Event}:
    \begin{itemize}
        \item Abnormal Period: Wednesday, January 22.
        \item Abnormal Type: Worked until 23:50 due to handling customer urgent needs, resulting in severe late bedtime.
        \item Impact on Health and Exercise: Directly led to no exercise on the same day, severely insufficient sleep duration (estimated < 6 hours), and relied on coffee to refresh mental state the next day (January 23), but boosted mental state through morning running.
        \item Recovery Measures and Effects: Went to bed before 22:30 on the evening of January 23 for sleep compensation, and recovered on January 24.
    \end{itemize}
\end{itemize}
\end{tcolorbox}

\subsection{QA Schema}
\begin{lstlisting}
{
    "question": "During the preparation for the daughter's graduation ceremony in June, on which day did the whole family (including the maternal grandmother, husband and son) participate in the final rehearsal or practice activity before the graduation ceremony?",
    "options": [
      {
        "option": "A",
        "content": "2025-06-22"
      },
      {
        "option": "B",
        "content": "2025-06-25"
      },
      {
        "option": "C",
        "content": "2025-06-28"
      },
      {
        "option": "D",
        "content": "2025-06-29"
      }
    ],
    "answer": "C",
    "required_events_id": [
      "2580"
    ],
    "required_events_date": [
      "2025-06-28"
    ],
    "ask_time": "2025-07",
    "question_type": "temporal",
    "evidence": [
      {
        "type": "agent_chat",
        "id": 409
      },
      {
        "type": "sms",
        "id": 1078
      }
    ]
}
\end{lstlisting}

\begin{lstlisting}
{
    "question": "What is the basis for Yu Xiaowei's proposal of the budget adjustment plan at the family video conference on June 15, 2025?",
    "answer": "It is based on the data of Chen Meiling's medical, medication and caregiver expenses in the past three months, as well as the key points of the optimized car insurance premium case she compiled. These data helped her present the family's financial situation and serve as specific cases and starting points for discussing financial optimization.",
    "score_points": [
      {
        "description": "Based on the data of the mother's medical, medication and caregiver expenses in the past three months",
        "score": 5
      },
      {
        "description": "Referring to the key points of the optimized car insurance premium case as specific cases and starting points for discussing family financial optimization",
        "score": 5
      }
    ],
    "required_events_id": [
      "2396",
      "2398",
      "2402"
    ],
    "required_events_date": [
      "2025-06-15"
    ],
    "ask_time": "2025-07",
    "question_type": "mutihop reasoning",
    "evidence": [
      {
        "type": "push",
        "id": 1182
      },
      {
        "type": "calendar",
        "id": "130"
      },
      {
        "type": "note",
        "id": "185"
      },
      {
        "type": "photo",
        "id": 877
      },
      {
        "type": "push",
        "id": 1183
      },
      {
        "type": "sms",
        "id": 983
      },
      {
        "type": "calendar",
        "id": "156"
      },
      {
        "type": "call",
        "id": 675
      },
      {
        "type": "note",
        "id": "238"
      }
    ]
}
\end{lstlisting}

\subsection{Plot Schema}
\begin{lstlisting}
{
  "topic": "The Road to Promotion: The Leap from Individual Performance to Team Management",
  "detailed_description": "This topic focuses on Xiaowei's journey to achieving her core career goal being promoted to Regional Sales Manager. The timeline starts in April 2025, where she accumulates key capital for promotion through the success of a landmark performance project. In the third quarter of 2025, under sustained performance pressure and with her contributions to team collaboration, she successfully passed the assessment and achieved a career leap within three years. After the promotion, she faced the challenge of transitioning from an individual contributor to a manager, taking on the role of a mentor for new team members for the first time and experiencing a difficult period of adaptation and integration. Relying on her inherent sense of responsibility and pragmatic spirit, she ultimately guided the new employees to success, gained initial recognition from the team, and capped off this phase with the honor of attending the Annual Excellent Sales Summit.",
  "monthly_description": [
    {
      "month": "April",
      "content": "Yu Xiaowei collaborated closely with her colleague Su Lishan. Leveraging Su Lishan's connections in Shenzhen and her own solid professional follow-up, she successfully designed a cross-border group insurance plan for a start-up tech company operating in both Shenzhen and Hong Kong, securing a major annual order. This project stands as an excellent demonstration of her cross-regional collaboration capabilities and proficiency in handling complex business.",
      "impact": "This large policy significantly boosted Yu Xiaowei's quarterly performance and became a highlight in her promotion application materials. Meanwhile, the successful collaboration with Su Lishan strengthened her partnerships within the team and accumulated positive experience for her in managing colleague relationships in her subsequent managerial role.",
      "core_events": [
        "Collaborated with colleague Su Lishan to secure a large group insurance policy"
      ]
    },
    {
      "month": "May - August",
      "content": "After securing the major order, Yu Xiaowei continued to maintain her high-end client network and deliver consistent performance. At the same time, the comprehensive wealth planning project for high-net-worth clients, which she had been collaborating on with Li Guodong, a bank wealth management manager, for nearly a year, entered the final critical stage. She devoted substantial effort to in-depth communication with clients and coordinated service integration between AIA and the bank, demonstrating professional planning and resource integration capabilities that go beyond pure sales. Her supervisor Liang Zhenhua noticed her all-round performance and began to consciously give her more opportunities to showcase herself in team meetings, while privately encouraging her to prepare for promotion.",
      "impact": "Her consistent high-quality performance and the ongoing major cooperation project earned her recognition and strong recommendation from her supervisor, a key prerequisite for passing the promotion assessment. This period was a sprint phase for her career goals, bringing increased work pressure but a strong sense of purpose.",
      "core_events": []
    },
    {
      "month": "September",
      "content": "Early in the month, the important client asset planning project in collaboration with Li Guodong was successfully signed, a family wealth inheritance plan integrating insurance, trust and banking services with a substantial value. In the middle of the month, relying on this project and many other outstanding achievements, a stable client network, and the strong recommendation of her supervisor Liang Zhenhua, Yu Xiaowei officially passed the company assessment and was promoted to Regional Sales Manager, responsible for managing a small sales team.",
      "impact": "The success of the project was a key milestone for her promotion, greatly enhancing her professional reputation and internal influence. The realization of the promotion is an important step in her career planning, bringing a higher professional status, income expectations and brand-new managerial responsibilities, marking her role transition from an individual contributor to a manager.",
      "core_events": [
        "Completed the important client asset planning project in collaboration with Li Guodong",
        "Was promoted to Regional Sales Manager"
      ]
    },
    {
      "month": "October - November",
      "content": "After the promotion, Yu Xiaowei faced her first managerial challenge: the company assigned a fresh graduate to her for her to act as a mentor. Initially, she tried to impart all her years of accumulated sales skills, but found that the new employee struggled to absorb the knowledge quickly and lacked initiative in practical work. This took up a great deal of her time originally dedicated to serving high-end clients, leading to a slowdown in her short-term individual performance growth and causing her anxiety. With the guidance of her husband Zhang Weiming and suggestions from her supervisor Liang Zhenhua, she adjusted her approach, shifting from meticulous hands-on guidance to setting clear goals and frameworks, encouraging independent exploration by the new employee and conducting regular reviews. Utilizing the enthusiastic and responsible traits of her ESFJ personality, she not only provided work guidance but also cared for the new employee in daily life, helping them integrate into the team.",
      "impact": "The mentorship experience was a painful growing pain of transition, testing her patience and leadership style. Through adjustments, she gradually found a balance between management and personal business. Successfully developing the new employee also initially proved her leadership potential, enhanced team cohesion, and boosted her confidence in her new managerial position.",
      "core_events": [
        "Took on the role of a mentor for new team members for the first time"
      ]
    },
    {
      "month": "December",
      "content": "Based on her annual target-accomplishing performance, Yu Xiaowei was invited to attend the company's Annual Excellent Sales Summit held in Hong Kong. She attended the summit with her supervisor Liang Zhenhua and her colleague and close friend Su Lishan. At the summit, she participated in high-end training and briefly shared a case in the sharing session where she led the new employee to collaboratively complete a small insurance policy.",
      "impact": "Attending the summit is the company's recognition of her work achievements over the past year, greatly enhancing her professional sense of honor and sense of belonging to the company. The new connections established at the summit may bring referral opportunities for the team's business in the future. It also draws a perfect close to the theme of her 2025 career promotion.",
      "core_events": [
        "Attended the company's Annual Excellent Sales Summit (Hong Kong Session)"
      ]
    }
  ]
}
\end{lstlisting}

\end{document}